\documentclass[pdflatex,sn-mathphys-num]{sn-jnl}

\usepackage{graphicx}%
\usepackage{multirow}%
\usepackage{amsmath,amssymb,amsfonts}%
\usepackage{amsthm}%
\usepackage{mathrsfs}%
\usepackage[title]{appendix}%
\usepackage{xcolor}%
\usepackage{textcomp}%
\usepackage{manyfoot}%
\usepackage{booktabs}%
\usepackage{algorithm}%
\usepackage{algorithmicx}%
\usepackage{algpseudocode}%
\usepackage{listings}%
\usepackage{comment} 

\usepackage[numbers]{natbib}

\theoremstyle{thmstyleone}%
\theoremstyle{thmstyletwo}%

\theoremstyle{thmstylethree}%

\begin{document}

\title[X Hedging]{X Hedging: towards an eXplainable artificial intelligence option Hedging framework}


\author[1]{\fnm{Bjørn André} \sur{Aaslund}}

\author[1]{\fnm{Johannes} \sur{Berge}}

\author[2]{\fnm{Ayoub} \sur{Haida}}

\author*[2]{\fnm{Ying} \sur{Ni}}\email{ying.ni@mdu.se}

\author[1]{\fnm{Rita} \sur{Pimentel}}

\affil[1]{\orgdiv{Department of Industrial Economics and Technology Management},
\orgname{Norwegian University of Science and Technology},
\orgaddress{\city{Trondheim}, \country{Norway}}}

\affil[2]{\orgdiv{Department of Business and Mathematics},
\orgname{M{\"a}lardalens University},
\orgaddress{\city{V{\"a}ster{\aa}s}, \country{Sweden}}}


\abstract{We propose X Hedging, an alternative to the Deep Hedging framework that addresses computational and explainability limitations of neural-network-based hedging strategies. Deep Hedging leverages neural networks to learn flexible hedging policies but relies on opaque models and typically requires large training datasets, which can complicate interpretation and limit its effectiveness in data-constrained settings.
X Hedging takes a step toward addressing these limitations by replacing neural networks with tree ensemble methods, specifically LightGBM, to estimate hedging strategies. This design provides a computationally efficient alternative while preserving strong hedging performance and moving toward more user-centered explainability.
To integrate LightGBM into a reinforcement-learning-style hedging framework, we introduce an optimization approach that sequentially trains tree-ensemble models while introducing stochasticity through random training orders.
In controlled experiments, X Hedging achieves performance comparable to Deep Hedging while offering greater user-centered explainability, improved computational efficiency, and substantially higher robustness when training data are limited. The numerical results demonstrate that tree-ensemble methods provide an effective computational alternative to neural networks for data-driven dynamic hedging.}




\maketitle

\section{Introduction}
Hedging is an important problem in computational finance, where market
participants such as market makers and over-the-counter (OTC) option
writers mitigate the risk of short option positions by taking offsetting
exposures in hedging instruments. Classical approaches originate from
the Black--Scholes model \citep{BS}, which provides analytical solutions
for pricing and hedging vanilla European options. However, such solutions rely on restrictive assumptions, including frictionless markets. Extensions incorporating market frictions, such as transaction costs, have therefore received considerable attention \citep{leland, neuberger, davis, atkinson}. Nevertheless, analytical tractability becomes increasingly limited as more realistic market features and hedging objectives are introduced. This has motivated the development of data-driven approaches that learn hedging strategies directly from simulated or observed data. More broadly, AI and
reinforcement-learning methods have been increasingly applied across financial risk management, asset pricing, return prediction, and related
financial problems \citep{aziz2019machine, chen2024deep, zhou2024much, RLHistory,hambly2023recent}. In option hedging specifically, this data-driven shift is reflected in a growing literature on neural-network-based methods \citep{reviewNN, li2024application, su2025deep, gnoatto2025deep}.

A prominent data-driven approach is \textit{Deep Hedging}, introduced
by \citet{deephedging}, which combines neural networks and reinforcement learning and has become a widely used framework for practical hedging applications \citep{RobustDeepHedge, DeepHedgeLongTerm, DeepHedgeRL, carbonneau2021equal, DeepHedgeEmpirical}.
Deep hedging, developed by researchers at JP Morgan, eliminates the need for strict assumptions, traditional Greeks, or specific pricing models, making it versatile for high-dimensional derivatives, various market dynamics, and frictions. The new hedging framework proposed in this work is inspired by Deep Hedging, retaining its dynamic hedging structure and the incorporation of transaction costs while replacing neural networks with tree ensemble methods. We coin the term \textit{X Hedging} for this new framework.

Deep Hedging leverages neural networks to achieve hedging policies with strong empirical hedging performance. While neural networks are highly flexible universal function approximators \citep{HORNIK1989359}, their internal decision logic is typically difficult to understand, offering limited insight into the structure of the learned hedging policy. This motivates the search for alternative computational approaches that retain the flexibility and hedging performance of data-driven methods while providing a higher degree of explainability. X Hedging is developed to address this problem.

This leads to the core question that guides our research:

\emph{Is it possible to design a data-driven hedging strategy that enhances model explainability without compromising hedging performance compared with the Deep Hedging approach?}

The area of research that focuses on developing methods and techniques to make AI system decision-making processes explainable and understandable is called Explainable AI (XAI). In finance, XAI has been explored across a wide range of applications \citep{chen2023explainable, weber2024applications}. The importance of explainability in financial applications is also recognized in policy-oriented literature. The Bank for International Settlements (BIS) \citep{bis}, as well as international AI principles developed by the OECD \citep{oecd2019principles} and G20 \citep{g20_2019_statement}, emphasize transparency and explainability as important elements of trustworthy AI systems.

Addressing explainability raises the question of how the concept should be understood, an issue widely debated in the literature \citep{schwalbe2024comprehensive}. In particular, the distinction between interpretability and explainability has drawn significant attention. While some authors treat the terms as interchangeable \citep{miller2019explanation}, others emphasize a conceptual distinction between them \citep{arrieta2020explainable, leblanc2023relationship}.
From a regulatory and practitioner perspective, however, these concepts are typically treated jointly as components of trustworthy AI. In recent reports on AI adoption by central banks \citep{BIS2025, BIS2025Explainability}, the BIS characterizes interpretable and explainable systems as those that provide sufficient information about their functionality and outputs to enable human users to understand and evaluate their behavior. 
This practical orientation motivates a user-centered view of explainability, in which the relevant users are human decision makers who must be able to understand and trust an AI system’s behavior. Consistent with this perspective, we adopt a user-centered notion of explainability throughout this work \citep{al2024user, laato2022explain, krakowski2025human}.

From a user-centered perspective, explainability is closely tied to model choice. A common distinction in the XAI literature is between inherently explainable models, whose structure is directly interpretable, and complex models that require post-hoc explanation methods to make their decisions intelligible \citep{lipton2018mythos}. 
Neural networks typically fall into the latter category, as their decision logic is opaque and must be explained ex post, whereas simpler models such as decision trees are often regarded as inherently explainable due to their transparent structure. In this work, we focus on developing a hedging model with intrinsic explainability, rather than relying on external post-hoc explanation mechanisms. This design choice aligns with recent calls for inherently explainable models in high-stakes settings \citep{rudin2019stop}, including the financial sector \citep{pimentel2025rethinking}.

X Hedging addresses the explainability limitations of Deep Hedging while retaining a similar overall structure, with the key distinction that tree-based ensemble methods replace neural networks in estimating hedging strategies. To the best of our knowledge, this is the first instance of such an approach being proposed in the literature. As noted by \citet{arrieta2020explainable}, tree ensemble methods do not qualify as fully explainable models in the strictest sense. Nevertheless, they retain a degree of structural comprehensibility because they are composed of inherently explainable elements---individual decision trees---and rely on rule-based decision logic that can, at least conceptually, be followed by human users. This more nuanced view is supported by \citet{herm2023stop}, which adopts a user-centered perspective on explainability and challenges the conventional dichotomy between performance and explainability. Their findings show that users' comprehension of tree ensemble models is nearly equivalent to their understanding of single decision trees, which the authors relate to similarities in algorithmic structure and the availability of tree-based visualizations. These findings motivate our use of tree ensembles as a computational alternative to neural networks that can preserve strong hedging performance while taking a step toward greater user-centered explainability.

As a tree ensemble model, we adopt LightGBM, introduced by \citet{lgbm}, which builds on the foundational Gradient Boosting Decision Trees (GBDT) framework \citep{GBDT}. LightGBM incorporates advancements from subsequent models, such as pGBRT \citep{tyree} and XGBoost \citep{xgboost}. Its primary distinction lies in its superior speed while maintaining comparable accuracy to these earlier methods. 
Building on its strong empirical performance in other domains, including retail sales forecasting \citep{MAKRIDAKIS2022}, cryptocurrency price trend prediction \citep{sun2020novel}, and financial default risk assessment \citep{gao2022construction}, we show that LightGBM is also a suitable and effective choice in the hedging context. 
At the same time, leveraging LightGBM within a hedging framework raises nontrivial methodological challenges, as tree-based models cannot be trained using stochastic gradient descent or backpropagation of the loss function. To address this, we propose a novel optimization approach for a reinforcement-learning-style hedging framework composed of multiple tree ensembles. 
Specifically, we fit all LightGBM models sequentially for one iteration, then train each model in a random order to introduce stochasticity. Furthermore, since LightGBM requires gradients and Hessians of the loss function, we construct a computational graph for the loss function to streamline the calculation process. 

Our results show that X Hedging achieves hedging performance comparable to Deep Hedging across various loss functions, both with and without transaction costs. This demonstrates that it is possible to design a hedging framework with greater user-centered explainability without compromising performance. 
In addition, X Hedging frequently exhibits faster computation times. In fact, it either outperforms or matches Deep Hedging in all experiments in terms of speed. 
Beyond explainability and efficiency, X Hedging also proves to be more robust to smaller training datasets. It maintains strong hedging performance even when the number of training paths is reduced by a factor of ten, whereas Deep Hedging performs poorly under the same conditions. 
This highlights an important advantage of X Hedging, which is its greater resilience to data limitations, a known shortcoming of Deep Hedging \citep{deephedging_vae}.

This paper contributes to the growing literature on computational methods for data-driven hedging by introducing a tree-based alternative to neural-network-based Deep Hedging. Our contribution is threefold. First, we introduce X Hedging, an AI-based hedging framework that replaces neural network policies with gradient-boosted tree ensembles, treating explainability as a design requirement rather than an afterthought. Second, we develop an optimization approach that enables the sequential training of tree-based policies within a reinforcement-learning-style hedging objective, addressing the absence of backpropagation in tree models. Third, using controlled simulation environments, we show that the resulting policies can match the hedging performance of Deep Hedging while improving explainability, computational efficiency, and robustness under limited data.

The rest of the paper is structured as follows. Section~\ref{sec:methodology} presents the X Hedging framework and the
relevant methods used in the numerical analysis.
Section~\ref{sec:performance_comparison} compares the performance of
X Hedging and Deep Hedging under the Black--Scholes setting, both with
and without transaction costs.
Section~\ref{Sec:CompaAdv} examines the comparative advantages of
X Hedging in terms of resilience to data constraints and user-centered
explainability.
Section~\ref{xhedging_with_jumps} evaluates the robustness of X Hedging
under Merton jump-diffusion dynamics.
Finally, Section~\ref{sec:conclusion} concludes the paper and outlines
potential directions for future research.

\section{X Hedging framework}
\label{sec:methodology}

X Hedging is a hedging framework in the spirit of the Deep Hedging model proposed by \citet{deephedging}.\footnote{The original development of X Hedging formed part of the master thesis by \citet{key}. The framework has subsequently been further developed in the present work.}

%
We start by defining a liability, $Z$, which may correspond to a European vanilla call or put option, with payoff $Z_T = \max \{ S_T - K, 0 \}$ or $Z_T = \max \{ K - S_T, 0 \}$, respectively, where $S_T$ is the value of the underlying asset at maturity $T$, and $K$ is the strike price.

The main motivation for a market maker or OTC trader is to manage the risk of their total position through hedging, thereby enabling them to profit from the bid–ask spread. 
Our framework hedges an option by dynamically adjusting positions in the underlying asset based on observed price movements. In the Black–Scholes setting, this reduces to the classical delta-hedging strategy.

The hedging occurs at equidistant time steps, $t_k$, with $k = 0, 1, 2, \dots n$, where $t_0=0$, $t_n = T$, and the step size is $\Delta t$. 
At each time step, $t_k$, the goal is to find the optimal strategy, $\delta_k \in \mathbb{R}$, of the hedging instrument, $S_k$.  Denote \( S := (S_k)_{k=0,\ldots,n} \) as the underlying asset price process at discrete time points, $t_k$, with \( k = 0,\ldots,n \), and \( \delta := (\delta_k)_{k=0,\ldots,n-1} \) as the hedger’s holdings in the underlying asset, which may be adjusted over the discrete trading times, $t_k$, with \( k = 0,\ldots,n-1 \). 
The cumulative 
gain process generated by the hedging strategy
$\delta$ is defined as
 \begin{equation*}
     (\delta \cdot S)_T := \sum_{k=0}^{n-1} \delta_k \cdot (S_{k+1}-S_k).
         \label{eq:deltastra}
 \end{equation*}

One of the main advantages of a non-parametric hedging model is that it can account for market frictions such as transaction costs and liquidity constraints.  The market frictions’ impact depends on the change in the hedging strategies, and their total sum may be written generally as
\begin{equation*}
    C_T(S, \delta) := \sum_{k=0}^{n} c(S_k,\delta_k - \delta_{k-1}),
    \label{eq:marketfrictions}
\end{equation*}
where $c: \mathbb{R}^+ \times \mathbb{R} \rightarrow \mathbb{R}^+$ represents the market friction function, and $\delta_k - \delta_{k-1}$ is the change in the hedging strategy between two consecutive time steps. For notational convenience we set $\delta_{-1} = \delta_n = 0$. 

In our model, we include transaction costs as market friction. 
Following \citet{leland} and  \citet{davis}, we consider proportional transaction costs, i.e., 
\begin{equation*}
    c(s,d) := \kappa |d| s,
    \label{eq:pcost}
\end{equation*}
where $\kappa$ is a positive constant representing the percentage of the transaction value charged. 
This means the cost is proportional to both the change in the hedging strategy and the value of the hedging instrument. Thus, a higher hedging instrument value and greater strategy changes increase costs, while lower values and smaller changes reduce them.

For a given asset price process $S$, terminal payoff $Z_T$, initial premium $p$,
and hedging strategy $\delta$, we denote the terminal profit and loss by
$\text{P}\&\text{L}_T(S,\delta,Z_T,p)$. 
For simplicity, hereafter, we suppress the dependence
of the $\text{P}\&\text{L}_T$ on $(S,Z_T,p)$, and write simply $\text{P}\&\text{L}_T(\delta)$,  whenever no ambiguity arises. 
Thus, the terminal $\text{P\&L}_T$ of the hedger is defined as 
\begin{equation}
    \text{P}\&\text{L}_T(\delta) := -Z_T + p + (\delta \cdot S)_T - C_T(S, \delta),
    \label{eq:pnl}
\end{equation}
Negative values of $\text{P}\&\text{L}_T$ correspond to losses, whereas positive
values indicate gains.

The goal of the hedger is to minimize the expected value of a predefined loss function, $\ell$, associated with the terminal P\&L, i.e.,
\begin{equation}
    \pi := \inf_{\delta} \mathbb{E} [ \ell ( \text{P}\&\text{L}_T ( \delta) ].
    \label{eq:minimize}
\end{equation}
Here the expectation is taken with respect to the underlying probability measure governing the asset price process $S$ (and hence the induced distribution of $Z_T$).

The optimal choice of $\delta$ is based on the information available to the hedger until each time step $t_k$, denoted by $I_k$. Our framework assumes $I_k = S_k$, i.e., the information about the underlying asset value at time step $t_k$. 
We keep the X Hedging framework simple to facilitate clear comparisons with Deep Hedging. However, it remains general enough to accommodate various types of liabilities with well-defined payoffs, incorporate multiple hedging instruments, and integrate additional information sources, such as news sentiment analysis or trader opinions.

Deep Hedging proposed by \citet{deephedging} uses a set of deep neural networks to map the information set $\{ I_0, I_1, \dots, I_{n-1}\}$ to the hedging strategies $\{ \delta_0, \delta_1, \dots, \delta_{n-1}\}$. A feed-forward neural network $\delta_k := F_k(I_k; \theta_k)$ for each time step $t_k$ is a function of the information at time $t_k$ given the neural network parameters $\theta_k$. The objective of the neural networks at each time step $t_k$ is to optimize the parameters $\theta_k$ such that $\pi$ in Equation \eqref{eq:minimize} is obtained. This is achieved by calculating a loss for each input and using backpropagation and stochastic gradient descent (SGD) to update the weights (we refer to \citet{goodfellow} for an in-depth explanation of these key concepts of neural networks). 

\begin{figure}
\begin{center}
\begin{minipage}{140mm}
\includegraphics[width=140mm]{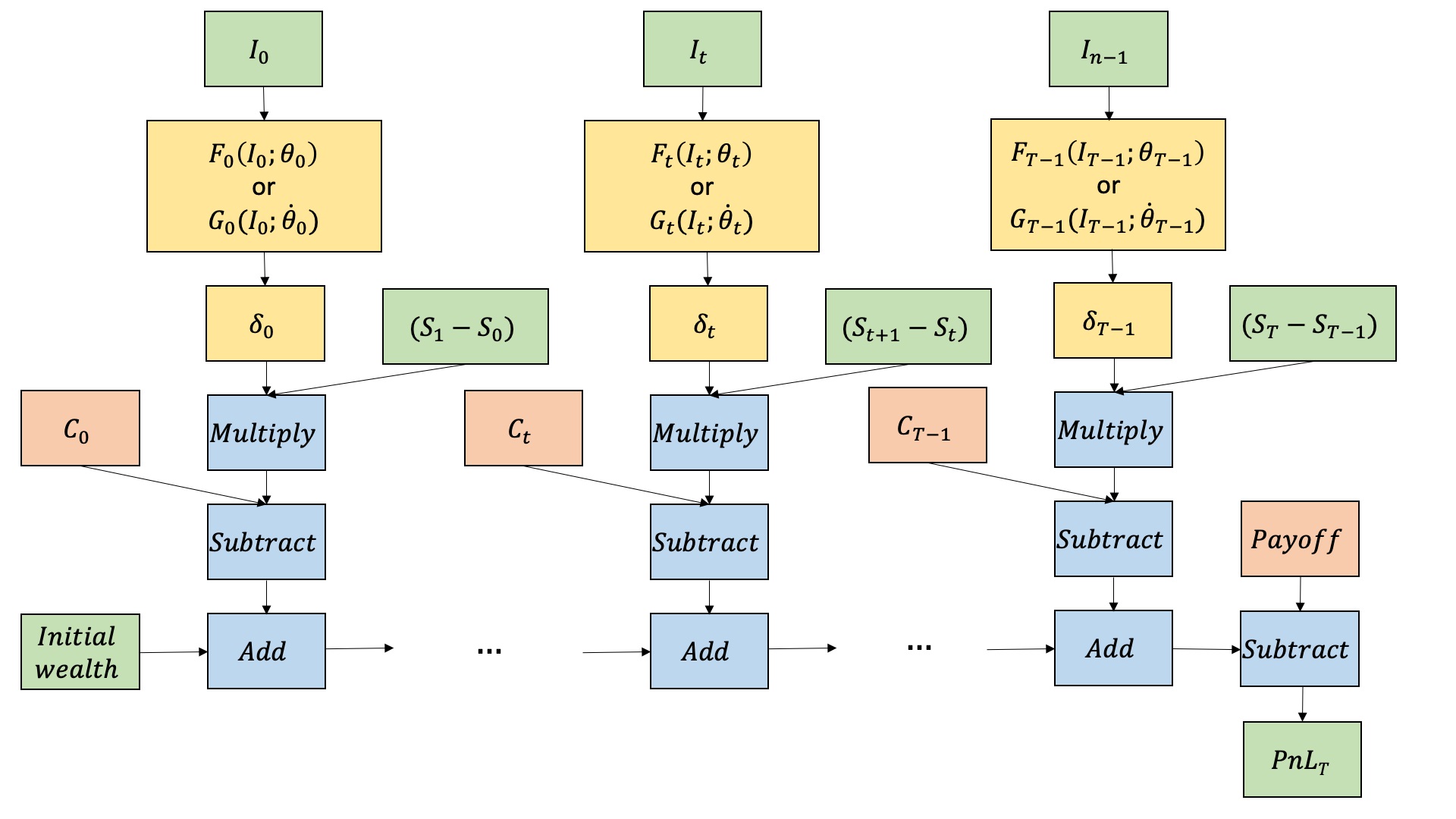}
\caption{Visual overview of X Hedging and Deep Hedging frameworks. Green boxes show inputs/outputs, blue boxes denote mathematical operations, orange boxes represent subtracted terms in Equation~\eqref{eq:pnl}, and yellow boxes indicate the neural network or tree-based model and its output.
}
\label{fig:model_overview}
\end{minipage}
\end{center}
\end{figure}

We propose fundamental modifications to Deep Hedging to enhance its user-centered explainability. Specifically, we suggest replacing the neural networks and their associated loss gradient calculations with tree ensemble methods, which have been considered more user-centered explainable models. Additionally, we introduce a custom loss function setup and optimization process. These changes form the foundation of X Hedging. 
We replace the neural networks $F_k(I_k; \theta_k)$ with a specific GBDT 
method $G_k(I_k; \dot{\theta}_k)$ for each time step $t_k$, where $\dot{\theta}_k$ are the parameters of the GBDTs. 
Figure \ref{fig:model_overview} shows both X Hedging and Deep Hedging for the first time step, an arbitrary time step, and the last time step. The X Hedging approach is described in Algorithm \ref{alg:dhlightgbm} in Appendix \ref{pseudocode}, and the code is available in this \href{https://github.com/HaidaQuant/Xhedging}  {repository}.

GBDT is a class of tree ensemble methods first proposed by \citet{GBDT}, and works by sequentially creating weak decision trees $g_{km}(I_k;  \ddot{\theta}_k)$ for $m \in \mathbb{N}_0$ where the model parameters $\ddot{\theta}_k$ are selected to improve the loss of the previous weak decision tree. This is called boosting, and the final ensemble is a linear combination of the weak decision trees, $G_k(I_k; \dot{\theta}_k) = \sum_m w_{km} g_{km}(I_k;  \ddot{\theta}_k)$, where $w_{km}$ is the importance weight associated with the $m$-th weak decision tree at time step $t_k$. An important attribute of the tree methods is that they provide explainability of each output by visually tracking every decision the model makes along the tree.

The specific GBDT method we implement is LightGBM, proposed by \citet{lgbm}. The key contribution of LightGBM is that it is more efficient than other GBDT models while still producing accurate results. \citet{lgbm} attributes this to their two novel techniques: Gradient-Based One-Side Sampling (GOSS) and Exclusive Feature Bundling (EFB). GOSS works by using all instances with large gradients and randomly sampling those with low gradients. This is an alternative to simply using a subset of the instances. EFB reduces the number of features by bundling the most important ones together. 

An important difference between X Hedging and Deep Hedging is that the former cannot use SGD and backpropagation of the loss function $\ell (\text{P\&L}_T)$ to obtain optimal values of $\dot{\theta}_k$ for Equation \eqref{eq:minimize}. This is due to the parameter update mechanism in LightGBM’s tree-based model. Instead, each LightGBM model $G_k$ must be fitted individually based on the global loss function $\ell (\text{P\&L}_T)$. 
After fitting all LightGBM models sequentially for one iteration, we train each model in a random order to introduce stochasticity. This randomness helps reduce the likelihood of converging to a local minimum rather than a global one, aligning with SGD's principles for avoiding local minima. To the best of our knowledge, this represents a novel approach to optimizing a reinforcement learning model composed of multiple tree ensemble models.

Compared to the neural networks, the LightGBM models need not only the gradients of the loss function but also the Hessians, the second derivative of the loss function $\ell (\text{P\&L}_T)$ with respect to $\delta_k$. The complexity of the loss function $\ell (\text{P\&L}_T)$ makes it difficult to derive a formula for the gradients and the Hessians analytically. Therefore, a computational graph of $\ell (\text{P\&L}_T)$ is created to simplify the calculation process. The graph uses automatic differentiation, specifically reverse-mode auto differentiation, for fast computation of the first and second-order derivatives, and then the chain rule is applied. 

LightGBM regressions are constructed by multiple trees, where each node in the trees divides the data based on a greater or equal condition for one feature. Therefore, the final regression line will consist of multiple orthogonal lines, making it non-smooth and noisy. However, the hedging strategies, $\delta_k$, are assumed to be smooth functions since the framework setup resembles the delta hedging by \citet{BS}. Based on this premise, a Savitzky–Golay filter, proposed by \citet{savitzky64}, is used to smooth the regression line. The smoothing is performed by a convolution, where successive adjacent data points are fitted to a second-order polynomial with linear least squares. This step is used solely for visualization purposes of the hedging strategies, i.e., it does not affect the performance of the results obtained from the $\text{P\&L}_T$ function.

The type of loss function, $\ell$, in Equation \eqref{eq:minimize} fundamentally determines the hedger's objectives. We explore two distinct loss functions, starting with the widely used mean squared error (MSE), defined as
\begin{equation*}
    \text{MSE}(Y) =\frac{1}{N} \sum_{i=1}^N (\hat{Y}_i - Y_i)^{2},
    \label{eq:mse}
\end{equation*}
where $N$ is the number of samples, $\hat{Y}_i$ are the calculated $\text{P\&L}_T$ in Equation \eqref{eq:minimize}, and $Y_i$ is set to zero to ensure that the bid-ask spread, not accounted for in Equation \eqref{eq:pnl}, becomes the hedger's profit.

The second loss function is based on the conditional tail moment (CTM) introduced by \citet{elmethni2014}. CTMs generalize the expected shortfall by considering higher-order moments of observations in the tail of the distribution and have recently also been studied in data-driven financial risk estimation \citep{allouche2025}. In particular, the first-order CTM corresponds to the expected shortfall, whereas the second-order CTM, denoted by $\mathrm{CTM}_2$, considers
the squared outcomes in the tail. This allows the optimization to focus specifically on hedging decisions associated with large adverse outcomes in the terminal $\mathrm{P\&L}_T$ distribution. Using terminal $\mathrm{P\&L}_T$, the empirical $\mathrm{CTM}_2$ is given by
\begin{equation}
    \widehat{\mathrm{CTM}}_{2,\alpha}
    =
    \frac{1}{|Y^{\prime}|}
    \sum_{y^{\prime}\in Y^{\prime}} (y^{\prime})^2,
    \label{eq:ctm2}
\end{equation}
where $Y^{\prime}$ denotes the set of the lowest $(1-\alpha)$-fraction of terminal $\mathrm{P\&L}_T$ outcomes, with $\alpha$ denoting the confidence level.

The use of $\mathrm{CTM}_2$ is particularly suitable for the
optimization in X Hedging. The training of the LightGBM models requires the gradients and Hessians of the objective loss function with respect to the hedging decisions. The squared tail outcomes in $\mathrm{CTM}_2$ provide the required non-zero Hessians.

In most of our experiments, we use the Black-Scholes model as a benchmark.  In these cases, we assume that, under the risk-neutral measure, the underlying asset, $S$, follows a Geometric Brownian Motion (GBM), 
\begin{equation*}
    S_t = S_0 \cdot \exp\left( \left( r - d - \frac{\sigma^2}{2}  \right) t + \sigma W_t \right),
    \label{eq:gbm-solution}
\end{equation*}
where $S_t$ is the underlying asset price at time $t$, $d$ is the continuous dividend yield, $\sigma$ is the annual volatility, $r$ is the constant interest rate, and $\{W_t: t\geq 0\}$ is a standard Brownian motion. 
The Black-Scholes model accounting for dividends, first presented by  \citet{merton1973}, can be used to solve the hedging strategies when there are no transaction costs. Indeed, to offset the risk produced by the movement of the underlying asset at time $t$, $\Delta_t$ underlying assets are purchased, where
\begin{align}
\Delta^{\text{call}}_t
  &= e^{-d (T-t)} N(d_1), \notag\\
\Delta^{\text{put}}_t
  &= e^{-d (T-t)} \bigl(N(d_1)-1\bigr),
\label{eq:deltabs}
\end{align}
with 
\begin{equation*}
d_1 = \frac{\text{ln}\left(\frac{S_t}{K}\right) + \left(r-d+ \frac{\sigma^2}{2} \right)(T-t)}{\sigma \sqrt{T-t}},    
\end{equation*}
and $N(\cdot)$ being the cumulative distribution function of a standard normal distribution. 
Replacing $\delta$ with $\Delta$ in Equation \eqref{eq:pnl} we get the baseline model $\text{P}\&\text{L}_T(S, \Delta, Z,p)$. 
%
The Black-Scholes delta hedging with proportional transaction costs, $\kappa$, can be obtained by discretizing the Black-Scholes equation and transforming the variance $\sigma^2$ into 
\begin{equation}
    \hat{\sigma}^2(\sigma^2,\kappa,\Delta t) = \sigma^2 \left[ 1+\kappa\frac{\sqrt{2/\pi}}{\sigma \sqrt{\Delta t}} \right].
    \label{eq:lelandsigma}
\end{equation}
This extended Black-Scholes model, proposed by \citet{leland}, is chosen for its simplicity and widespread adoption in the literature.

To test the X Hedging framework under extreme conditions, we also consider
the Merton jump-diffusion model \citep{merton1976option} for the underlying
asset under the risk-neutral measure,
\begin{equation}
\label{jump_model}
S_t = S_0 \cdot \exp\left( \left( r - d-r_J - \frac{\sigma^2}{2}  \right) t + \sigma W_t \right) \cdot \prod_{i=1}^{N_t} J_i,
\end{equation}
where $\{N_t, t\geq 0\}$ is a Poisson process with intensity $\lambda_J$ and \( J_i := e^{Y_i} \) represent the jumps, which are independent and identically distributed 
with log-jump sizes \( Y_i \sim \mathcal{N}(\mu_J, \sigma_J^2) \).  The risk-neutral drift correction for jumps is given by $r_J := \lambda_J (e^{\mu_J + \sigma_J ^2/2} -1)$.

Since the presence of jumps makes the market incomplete, no-arbitrage
alone does not uniquely determine the pricing measure and the corresponding
delta. We therefore use the delta associated with the risk-neutral Merton
specification in Equation~\eqref{jump_model} as a consistent model-based
benchmark. We compute this delta using the semi-analytical Fourier-based
formula (with no dividends) for the Greek $\Delta$ of a call option,
adapted from Chapter 5 of \citet{kienitz2012financial} and based on the
Fourier approach of \citet{lewis2001simple}.

\begin{align}
\label{delta_jump}
\Delta^{\text{call}}_{t}
&= 1 - \frac{K e^{-r(T-t)}}{S_t \pi}
\notag\\
&\quad \times
\int_0^{+\infty}
\frac{\Re\!\left[
(iu+\tfrac12)e^{(iu+\tfrac12)k}
\Phi_t\!\left(u-\tfrac{i}{2}\right)
\right]}{u^2+\tfrac14}
\,du,
\end{align}
where $k= log(S_t/K)$ and $\Re[x]$ denotes the real part of $x$. Here $\Phi_t(\cdot)$ is the characteristic function for the stock price in log terms, given as
\begin{align*}
\Phi_t(u)
&= \exp\Big(
(T-t)\,\Psi(u)
\Big), \\
\Psi(u)
&:= iu\,\omega - \tfrac12\sigma^2u^2
+ \lambda_J\!\left(
e^{iu\mu_J-\tfrac12\sigma_J^2u^2}-1
\right), \label{psi_def}
\end{align*}
with $\omega := r - \frac{1}{2} \sigma^2 - \lambda_J \left( e^{\mu_J + \frac{1}{2} \sigma_J^2} - 1 \right)$. 

\section{X Hedging Performance in Comparison with Deep Hedging}
\label{sec:performance_comparison}

We conduct numerical experiments for X Hedging and Deep Hedging on a PC with 1.3GHz Intel Core i5 CPU - 12 cores and 16Gb RAM. 
Unless stated otherwise, the hedging models are trained in-sample on $N=200,000$ simulated paths and tested out-of-sample on an independent set of $N/2$ paths. Some selected experiments use smaller values of $N$. 
Appendix \ref{app:parameters} lists the parameters used for the LightGBM models in X Hedging and the neural networks in Deep Hedging. 

We consider European vanilla call and put options with a time horizon of 30 trading days with daily rebalancing. Following \citet{deephedging}, we assume that all intermediate payments are accrued at a risk-free overnight rate, which can reasonably be considered negligible and thus set to zero.

Using two loss functions, namely MSE and  CTM$_2$, we validate the performance of the X Hedging framework, demonstrating that it produces accurate hedging strategies comparable to benchmark models. Moreover, across experiments comparing X Hedging and Deep Hedging, X Hedging outperforms in most cases in computational speed, achieving an average reduction of approximately 30\% in computation time.


We discuss the results where the underlying asset follows a GBM in Section \ref{sec:ValidationCVaR}, and a jump-diffusion in Section \ref{xhedging_with_jumps}.

\label{sec:ValidationCVaR}

We begin by validating the X Hedging (XH) framework on two baseline cases involving European at-the-money (ATM) call options, assuming the underlying asset follows a GBM and using the MSE loss function. The two scenarios considered are one without transaction costs and another with proportional transaction costs, using a fixed cost parameter of $\kappa = 0.001$. 
We compare the results with Deep Hedging (DH) and a benchmark. In the absence of transaction costs, the benchmark hedging strategy is the original Black-Scholes (BS), with delta given by Equation \eqref{eq:deltabs}. When transaction costs are included, we adopt the Leland Hedging strategy (BS-L), which adjusts the BS delta using a modified volatility parameter as given in Equation \eqref{eq:lelandsigma}. 
The results indicate that both X Hedging and Deep Hedging perform similarly to the benchmark strategies, with detailed numerical results provided in Appendices \ref{app:baseCase1} and \ref{app:baseCase2}. A noteworthy observation is that the terminal $\text{P\&L}_T$ is significantly more negative when transaction costs are included. This is evident from the comparison of distribution statistics in Tables \ref{tab:results-none-mse} and \ref{tab:results-proportional-mse}, and the histograms in Figures \ref{fig:histogram-none-mse} and \ref{fig:histogram-proportional-mse}. This outcome is consistent with expectations, as transaction costs naturally increase the overall cost of hedging.

Next, we consider the more comprehensive scenario with CTM$_2$ loss function defined in Equation~\eqref{eq:ctm2}, considering proportional transaction costs. A higher confidence level $\alpha$
focuses the loss function on a smaller fraction of more extreme adverse outcomes, but also reduces the number of observations contributing to the loss during optimization, which can slow convergence. Therefore, there is a trade-off between the emphasis on extreme tail losses and computational efficiency. We select $\alpha=0.5$ to improve convergence while still placing greater emphasis on adverse outcomes in the terminal $\mathrm{P\&L}_T$ distribution.

Table \ref{tab:parameters} presents the parameters used to simulate the underlying asset for ATM European call and put options. 

\begin{table}[h]
\caption{GBM model parameters for asset price simulation and option parameters.}
\label{tab:parameters}
\begin{tabular}{@{}cccccccc@{}}
\toprule
$S_0$ & $K$ & $r$ & $d$ & $\sigma$ & $T$ & $n$ & $N$ \\
\midrule
$1$ & $1$ & $0.0$ & $0.0$ & $0.3$ & $\frac{30}{365}$ & $30$ & $200000$ \\
\botrule
\end{tabular}
\end{table}

\begin{figure}[h]
\centering

\begin{minipage}[t]{0.48\textwidth}
\centering
\includegraphics[width=\linewidth]{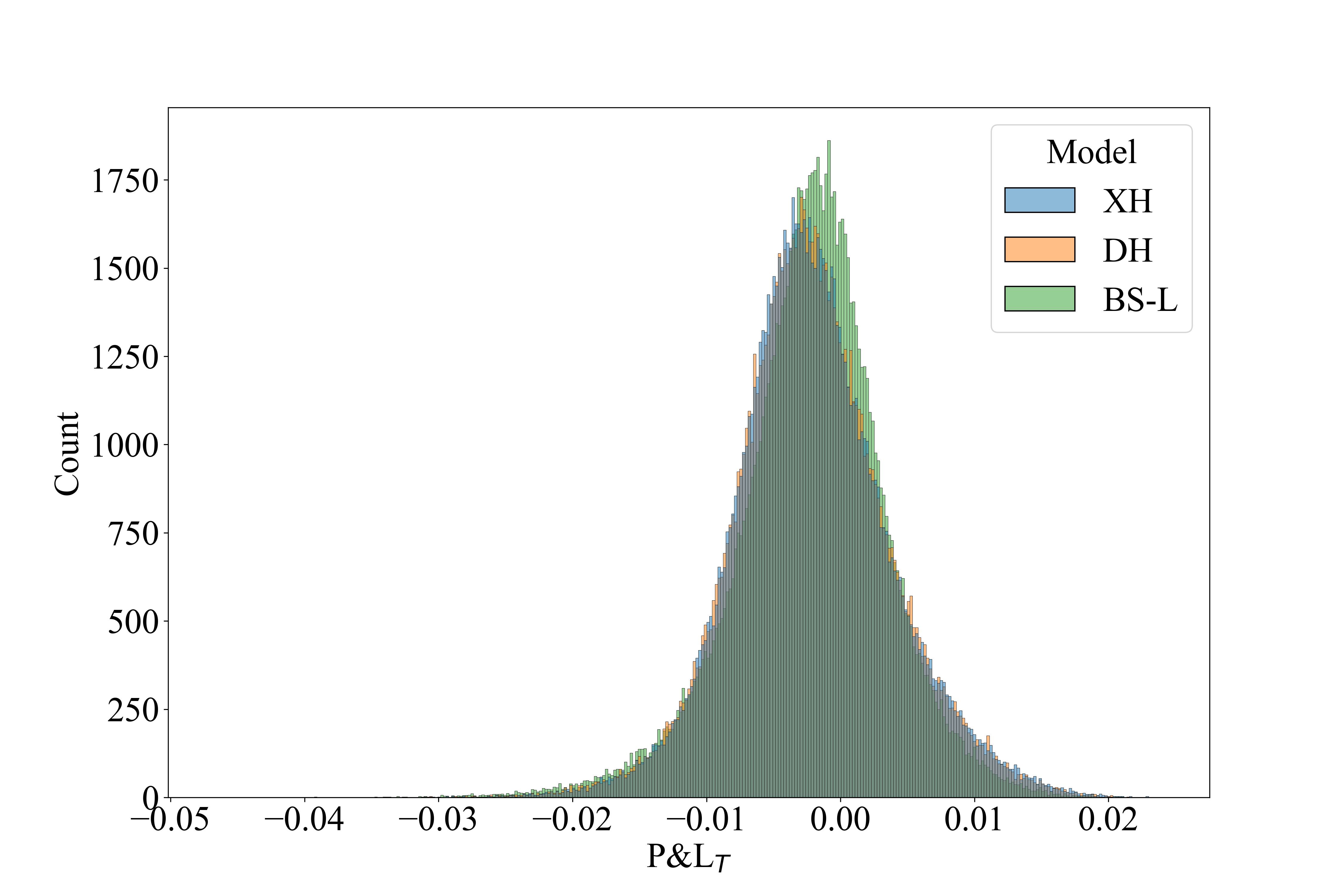}\\
(a) Call option.
\end{minipage}
\hfill
\begin{minipage}[t]{0.48\textwidth}
\centering
\includegraphics[width=\linewidth]{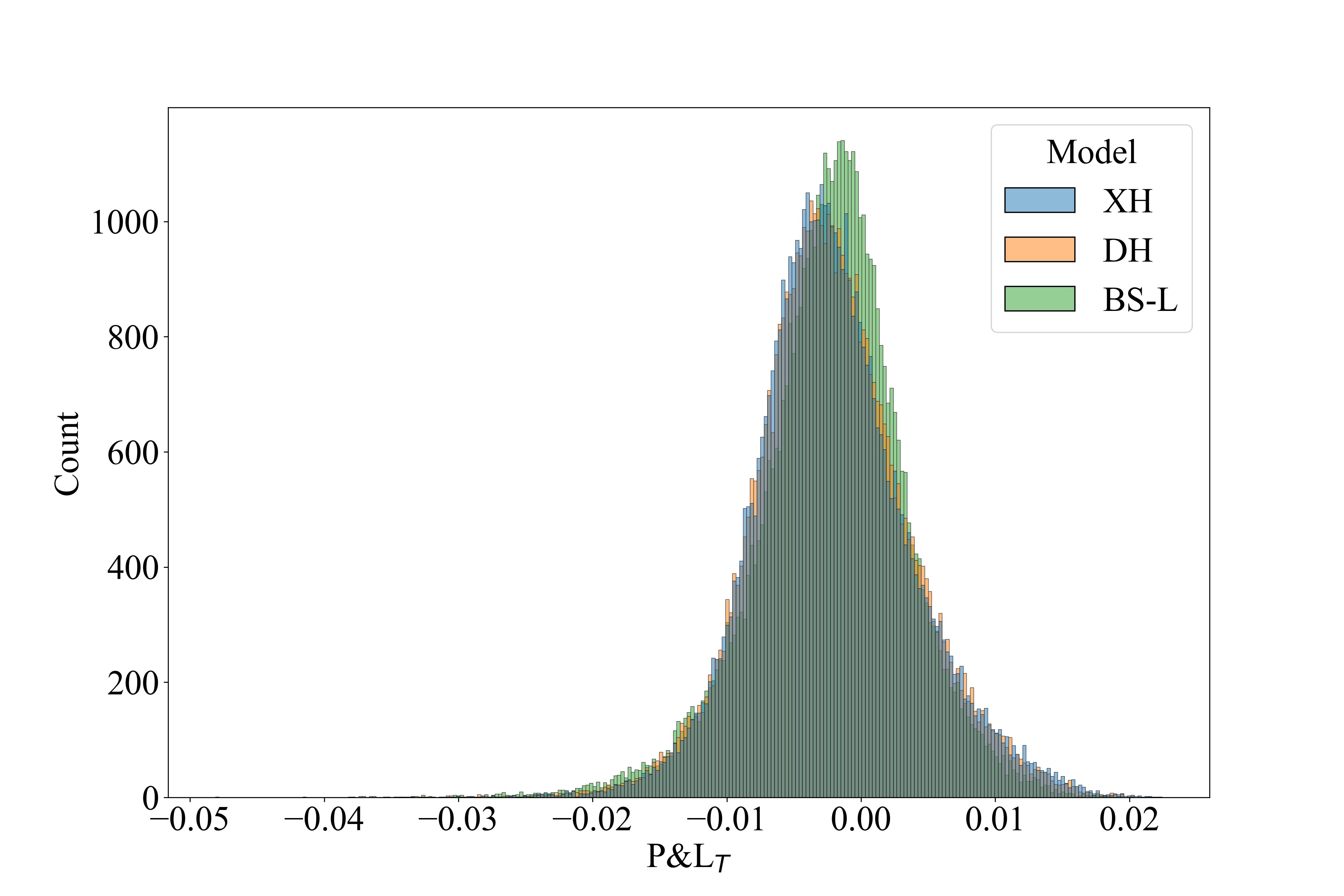}\\
(b) Put option.
\end{minipage}

\caption{Histograms of terminal $\text{P\&L}_T$ distributions for ATM options under transaction costs using the CTM$_2$ loss function.}
\label{fig:histogram-proportional-cvar-square}
\end{figure}

\begin{figure}
\begin{center}
\begin{minipage}{\textwidth}
\centering
\includegraphics[width=0.30\textwidth]{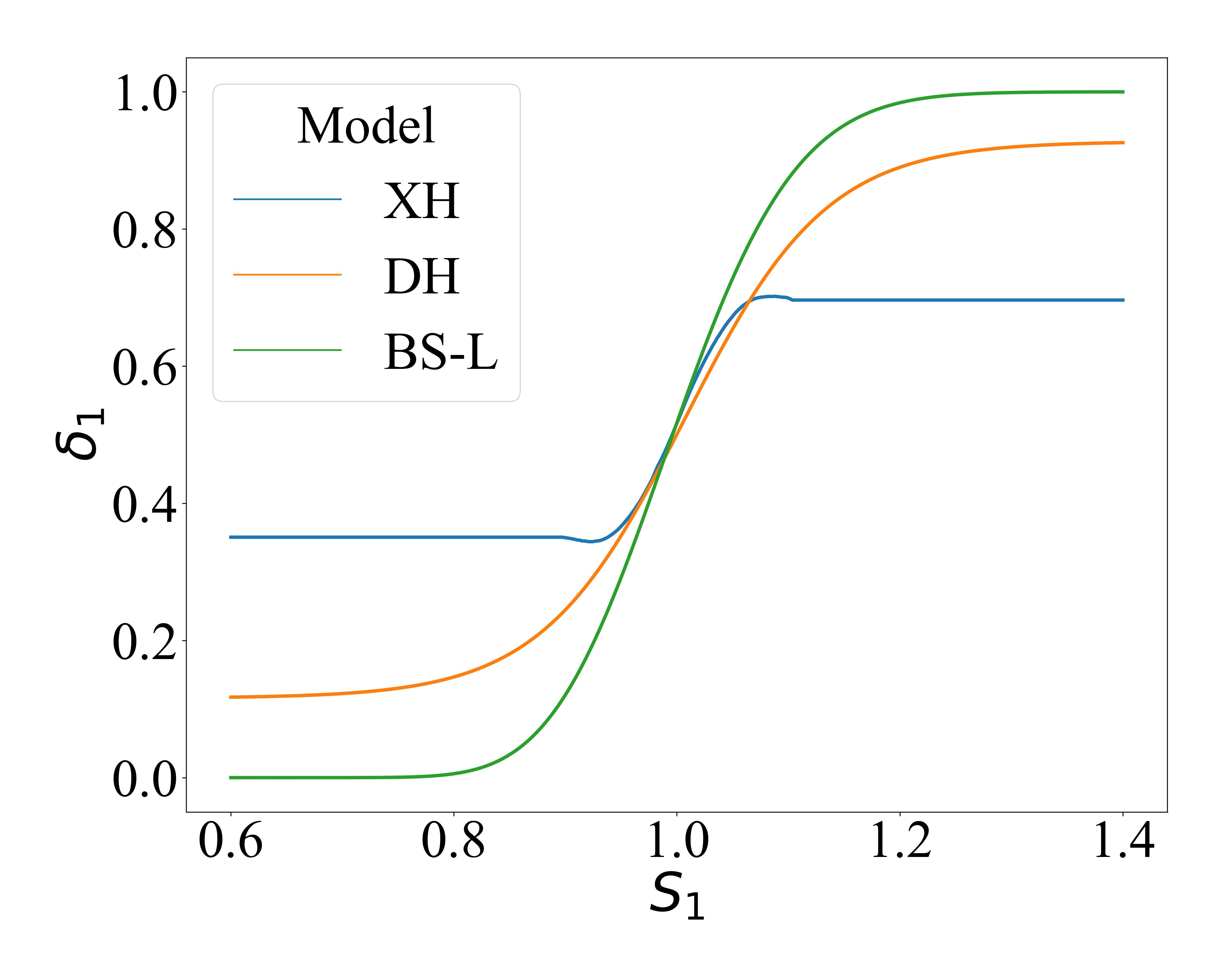}\hfill
\includegraphics[width=0.30\textwidth]{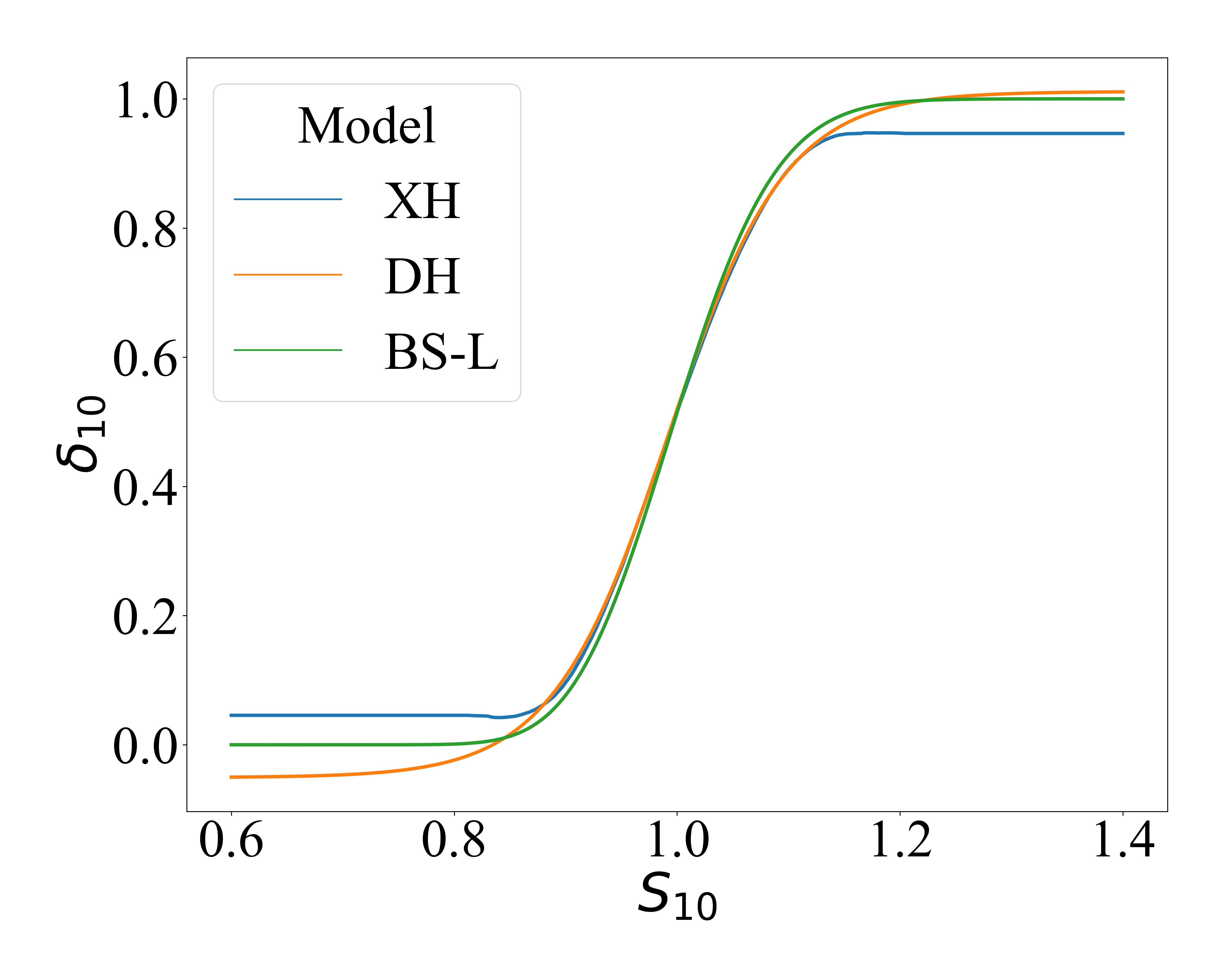}\hfill
\includegraphics[width=0.30\textwidth]{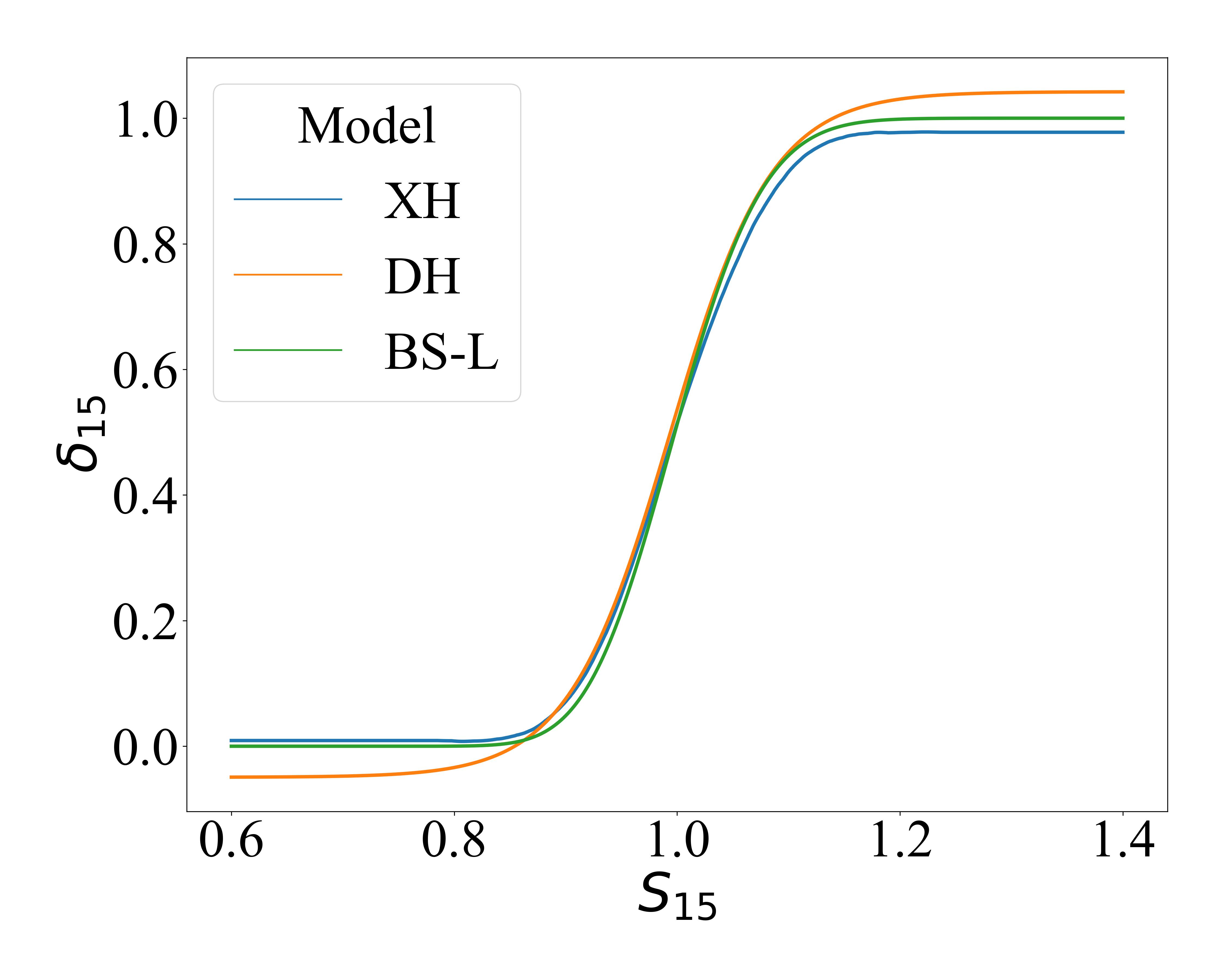}\\[-2pt]
\includegraphics[width=0.30\textwidth]{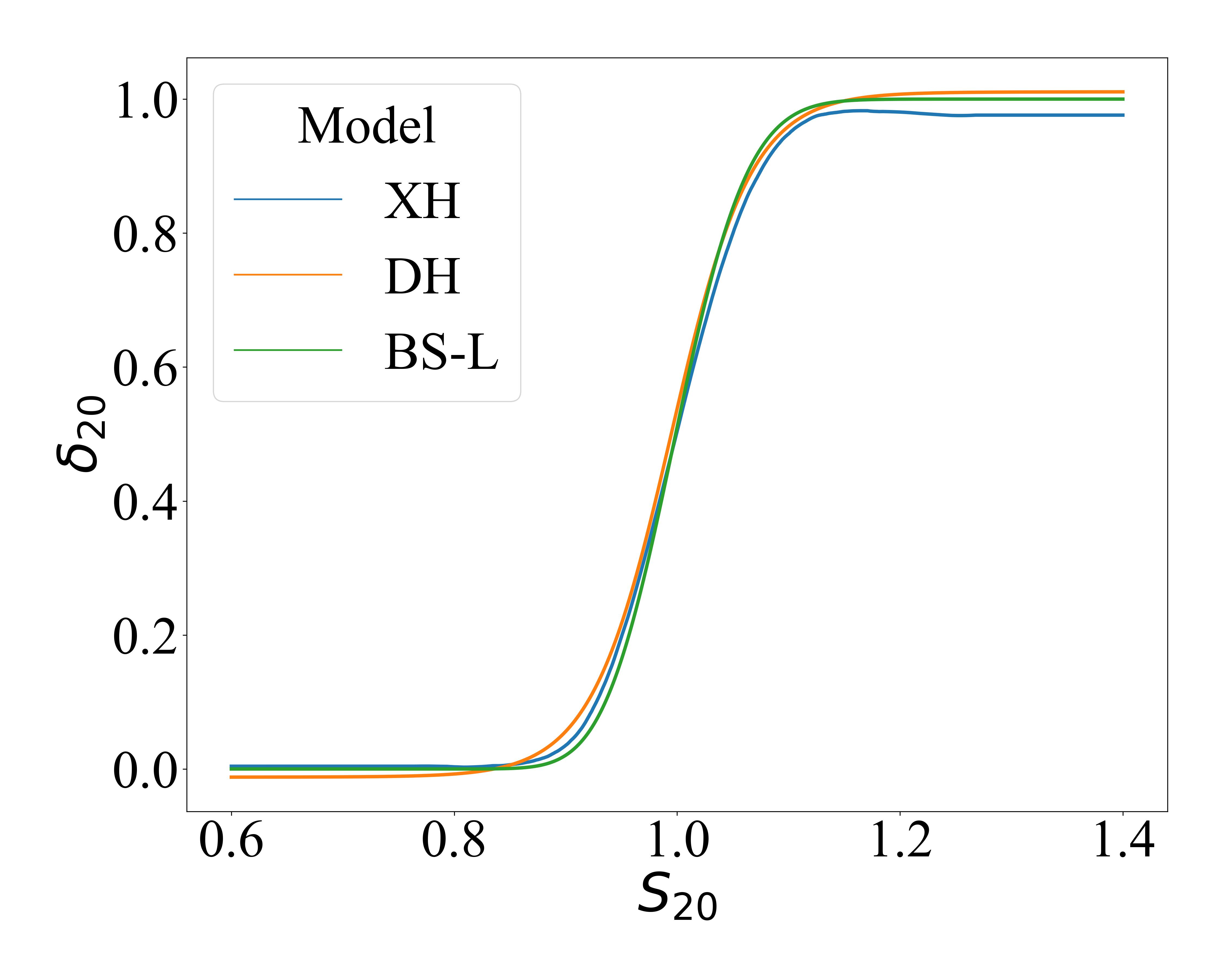}\hfill
\includegraphics[width=0.30\textwidth]{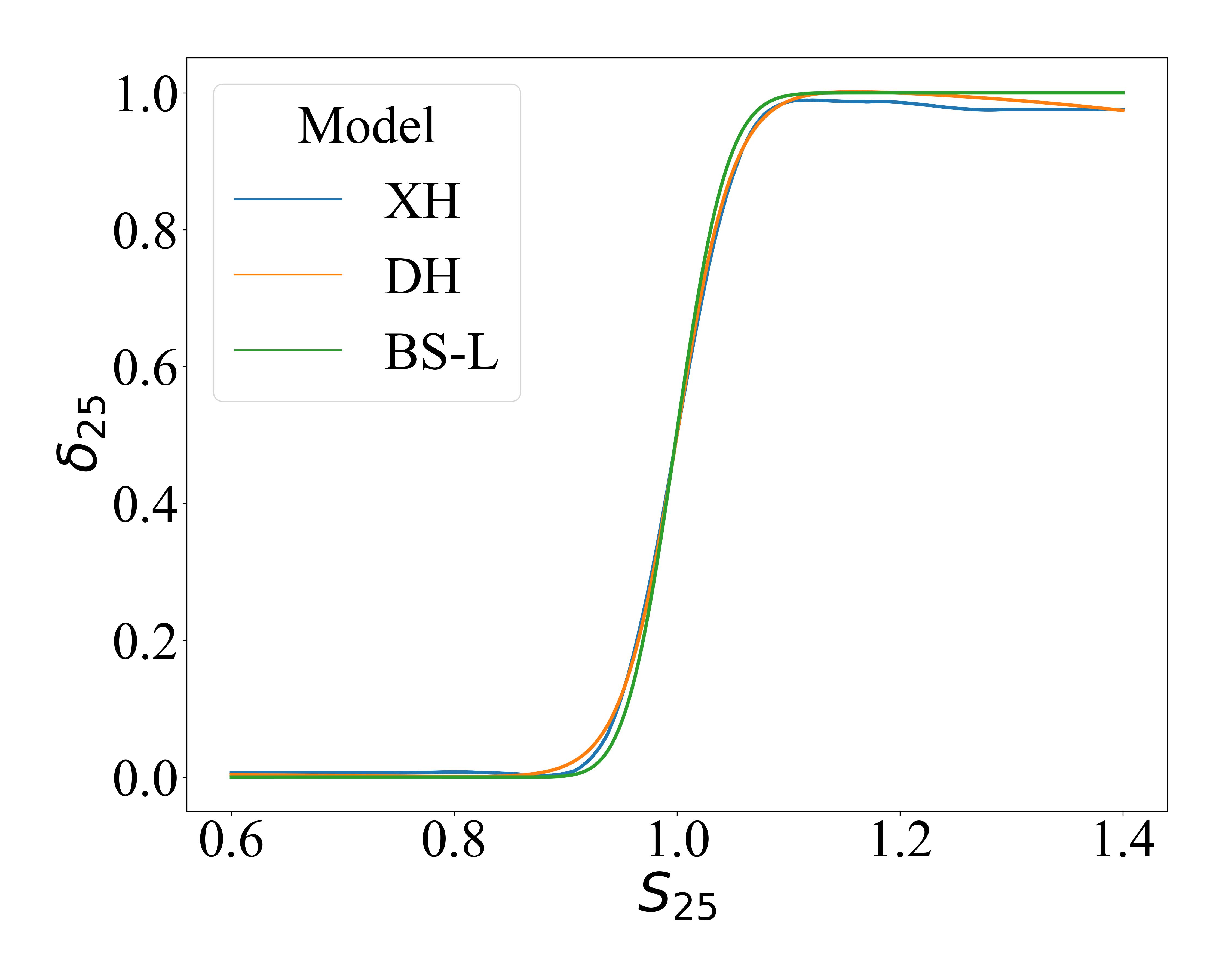}\hfill
\includegraphics[width=0.30\textwidth]{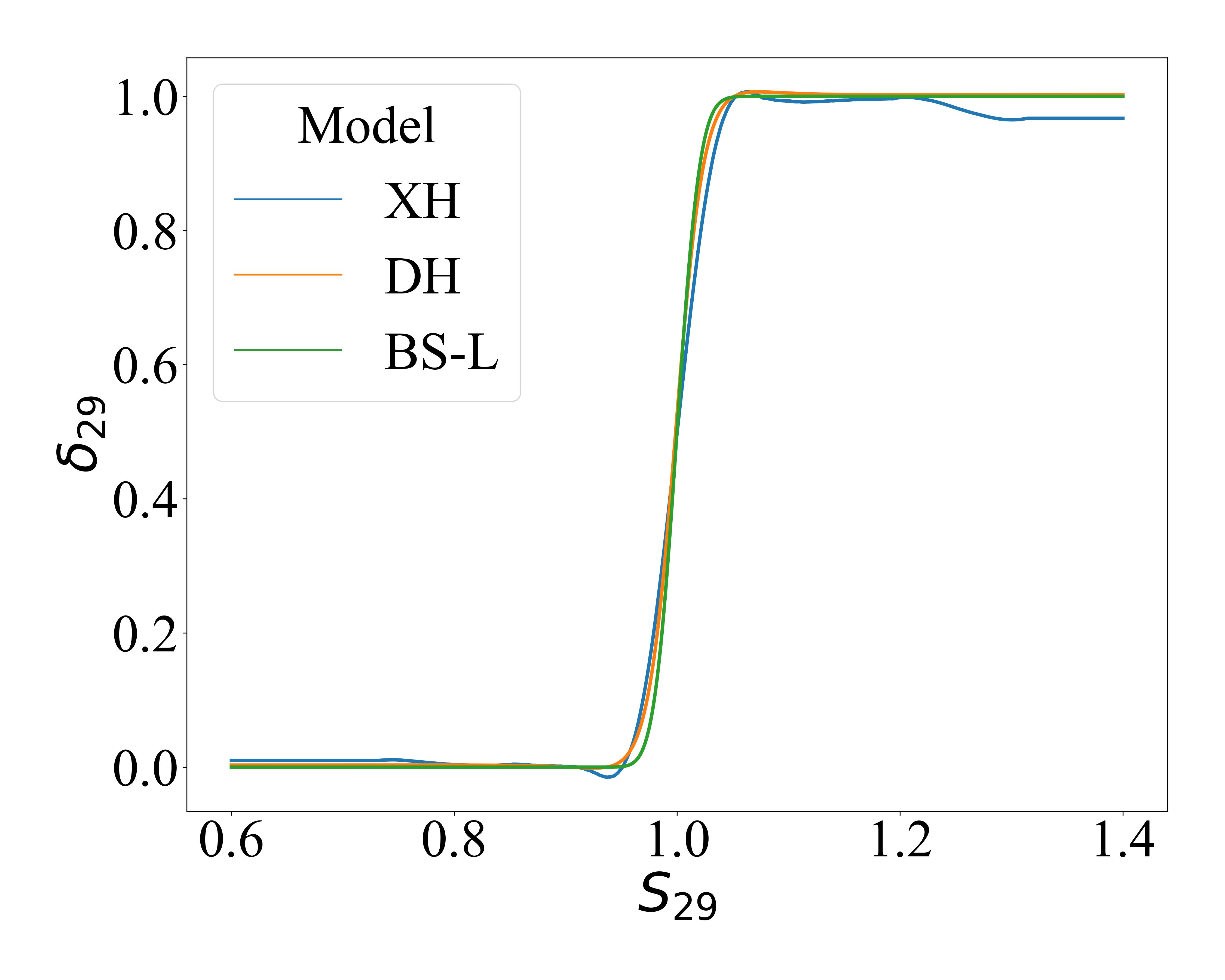}
\caption{Hedging strategies for ATM call option at times $k=1,10,15,20,25,29$ under transaction costs and the CTM$_2$ loss function.
\label{fig:strategy-proportional-cvar-square-call}}
\end{minipage}
\end{center}
\end{figure}

\begin{figure}
\begin{center}
\begin{minipage}{\textwidth}
\centering
\includegraphics[width=0.30\textwidth]{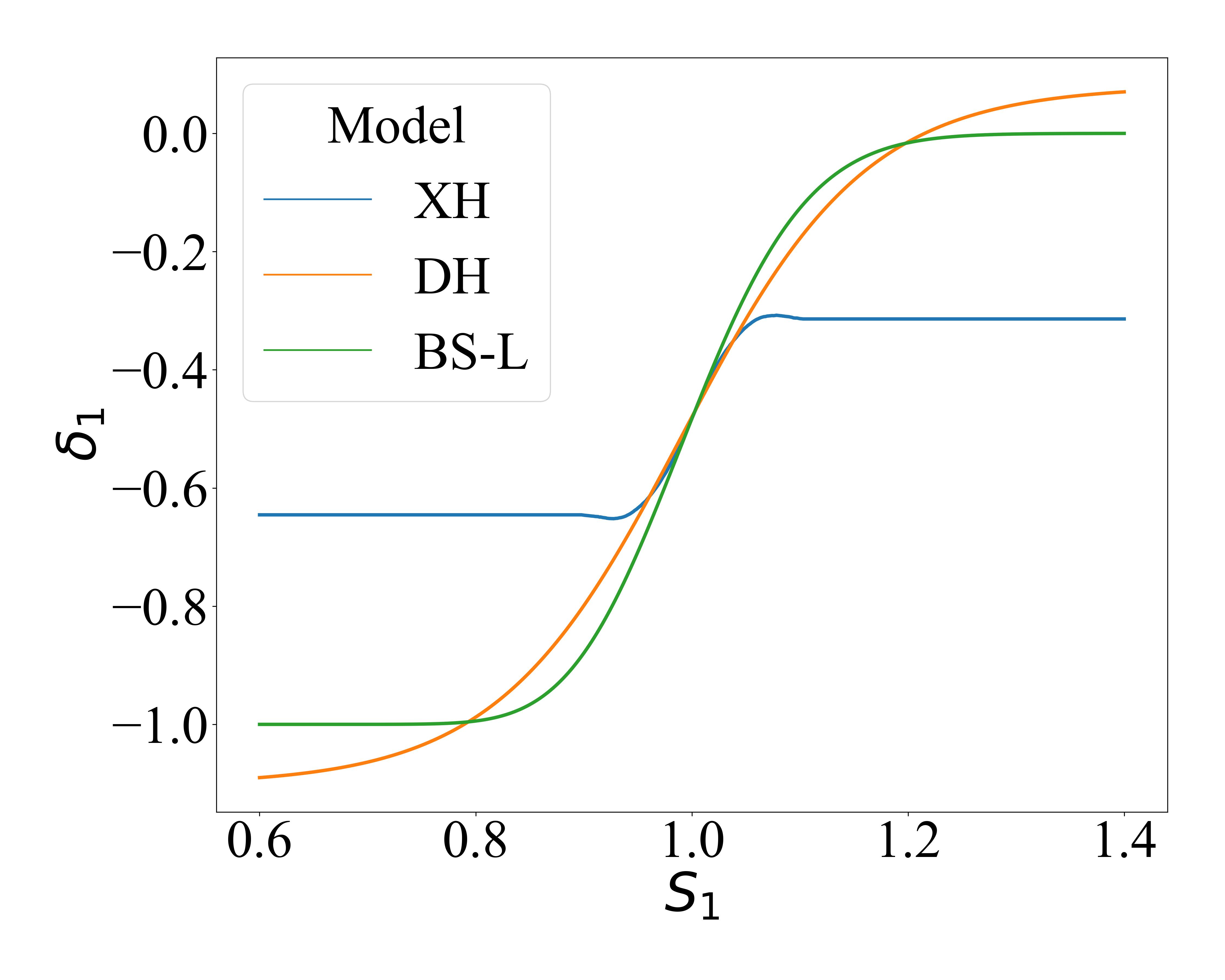}\hfill
\includegraphics[width=0.30\textwidth]{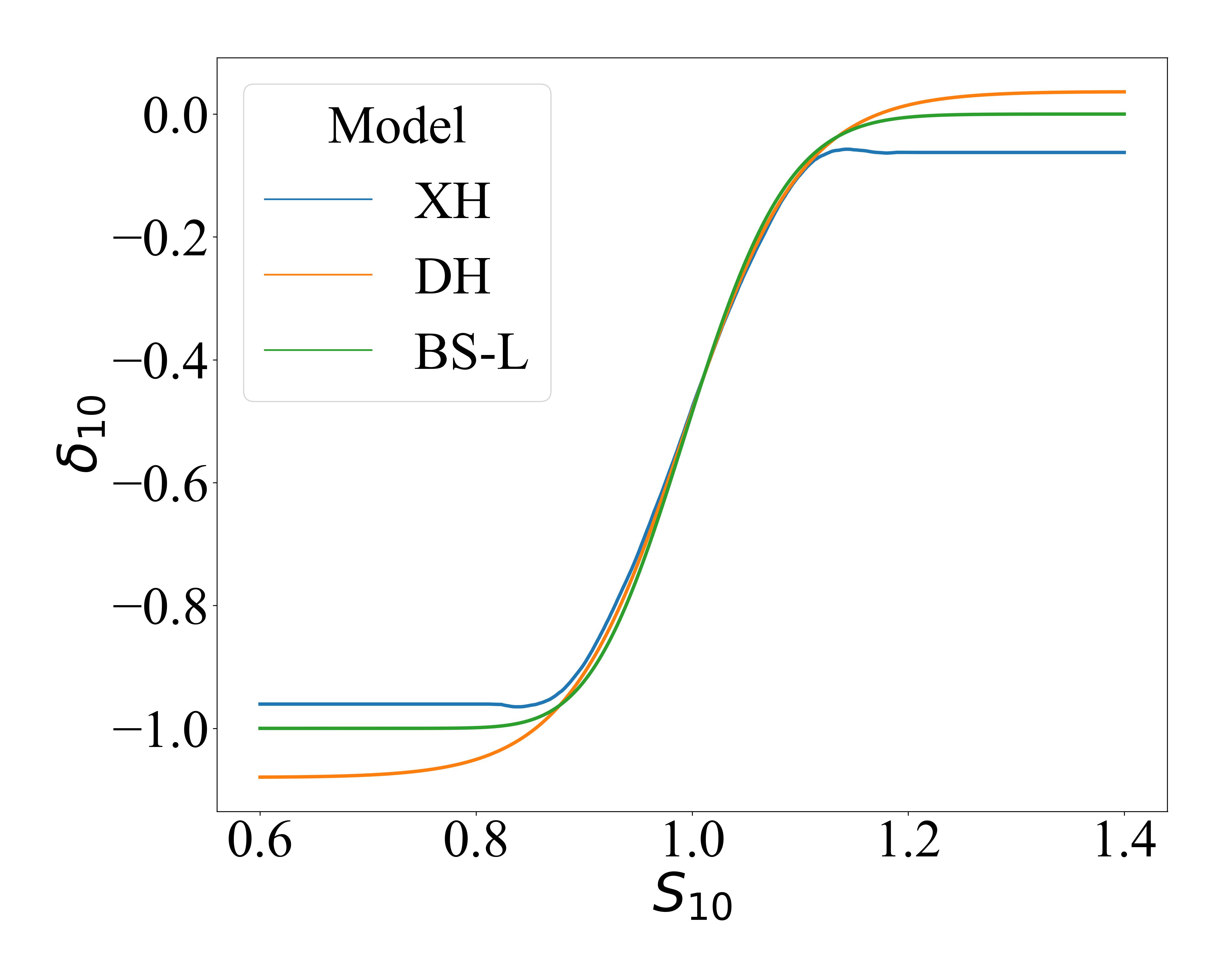}\hfill
\includegraphics[width=0.30\textwidth]{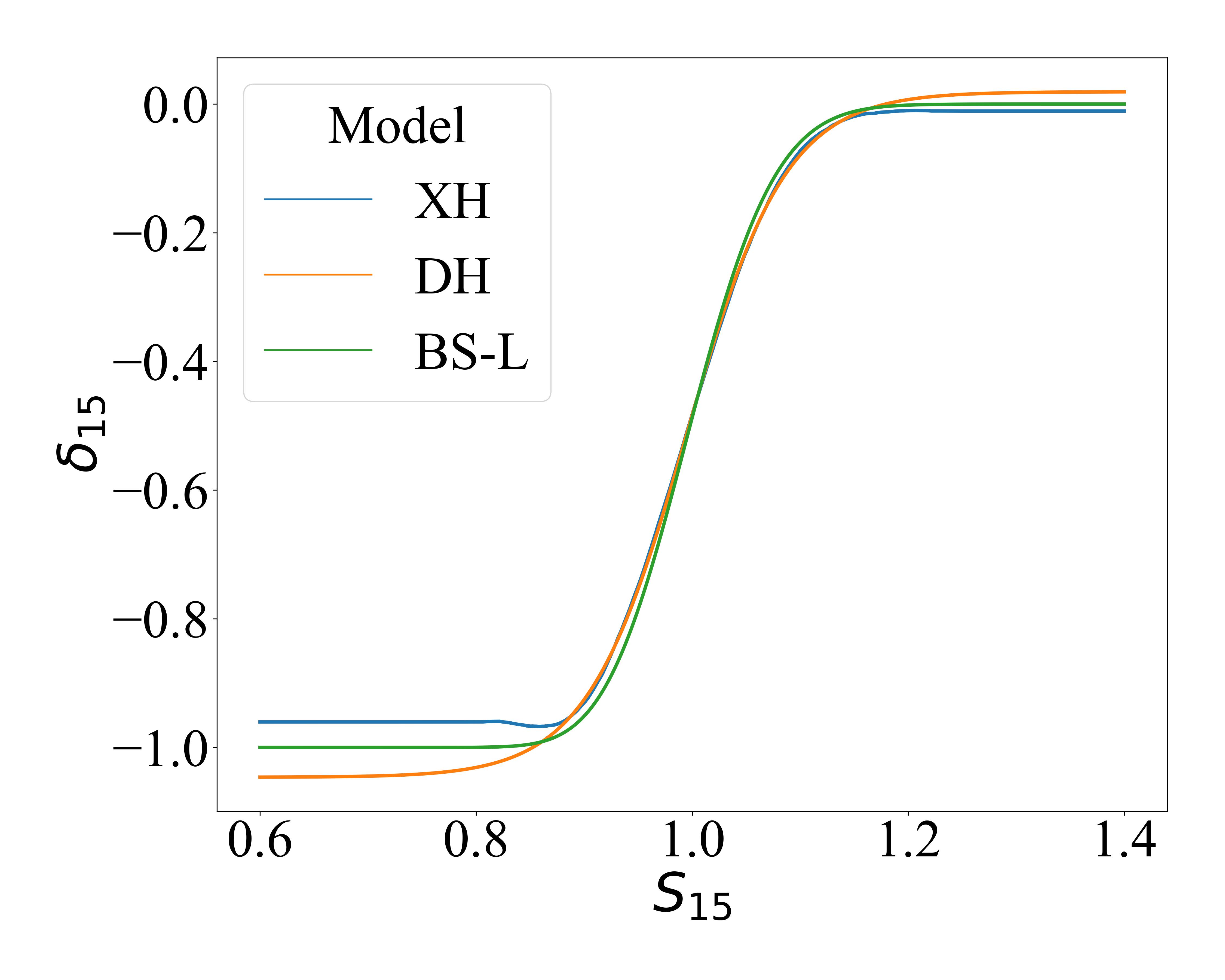}\\[-2pt]
\includegraphics[width=0.30\textwidth]{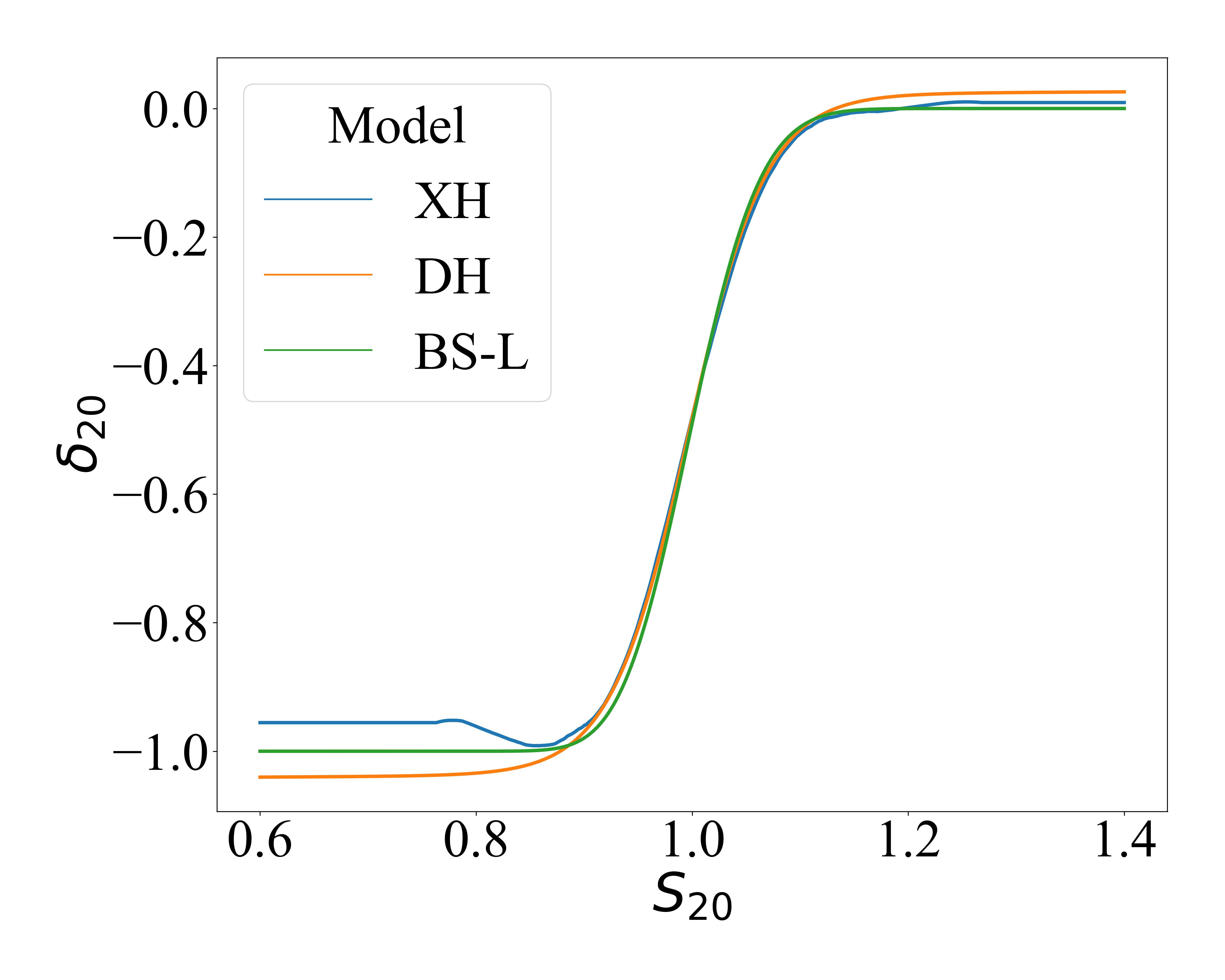}\hfill
\includegraphics[width=0.30\textwidth]{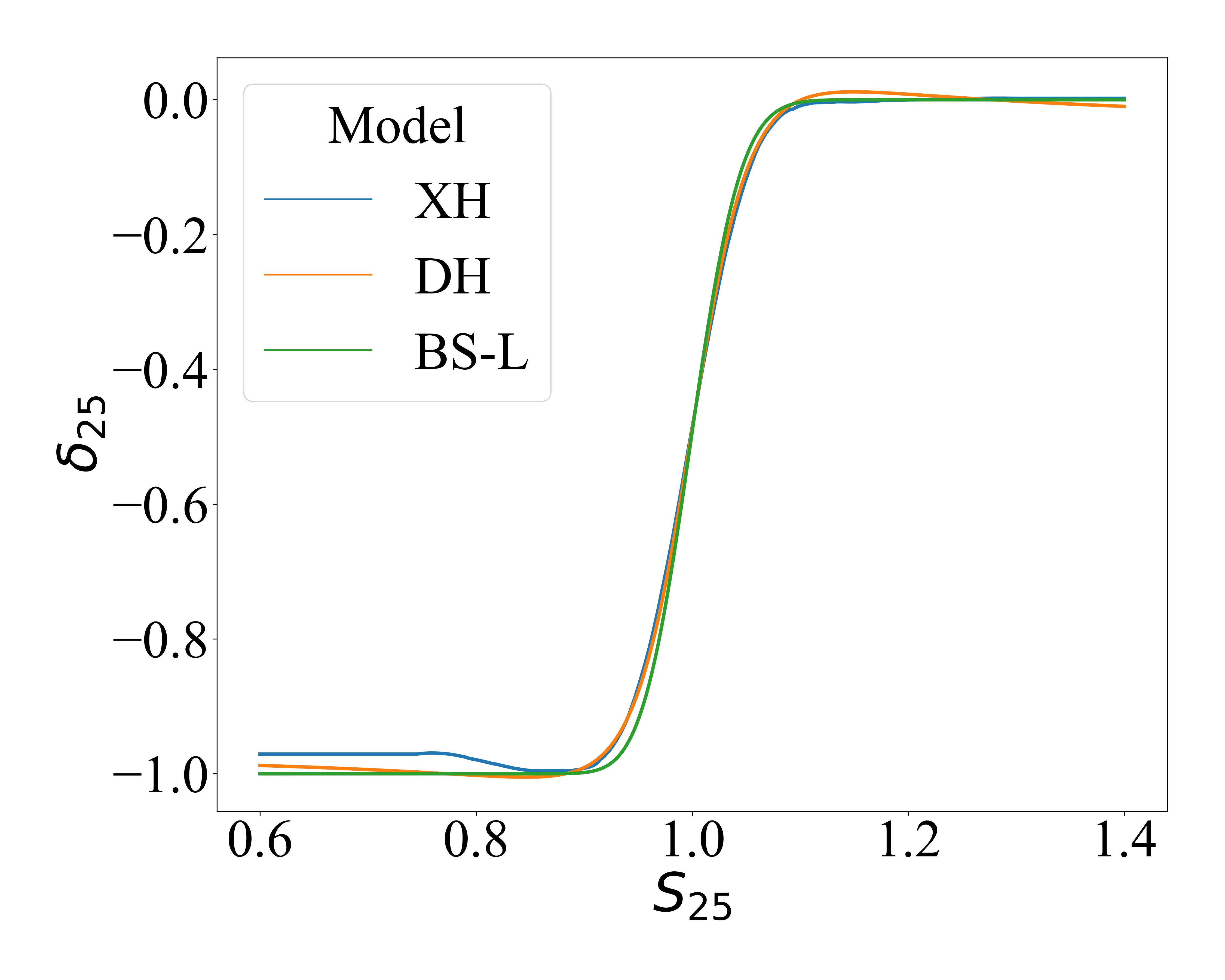}\hfill
\includegraphics[width=0.30\textwidth]{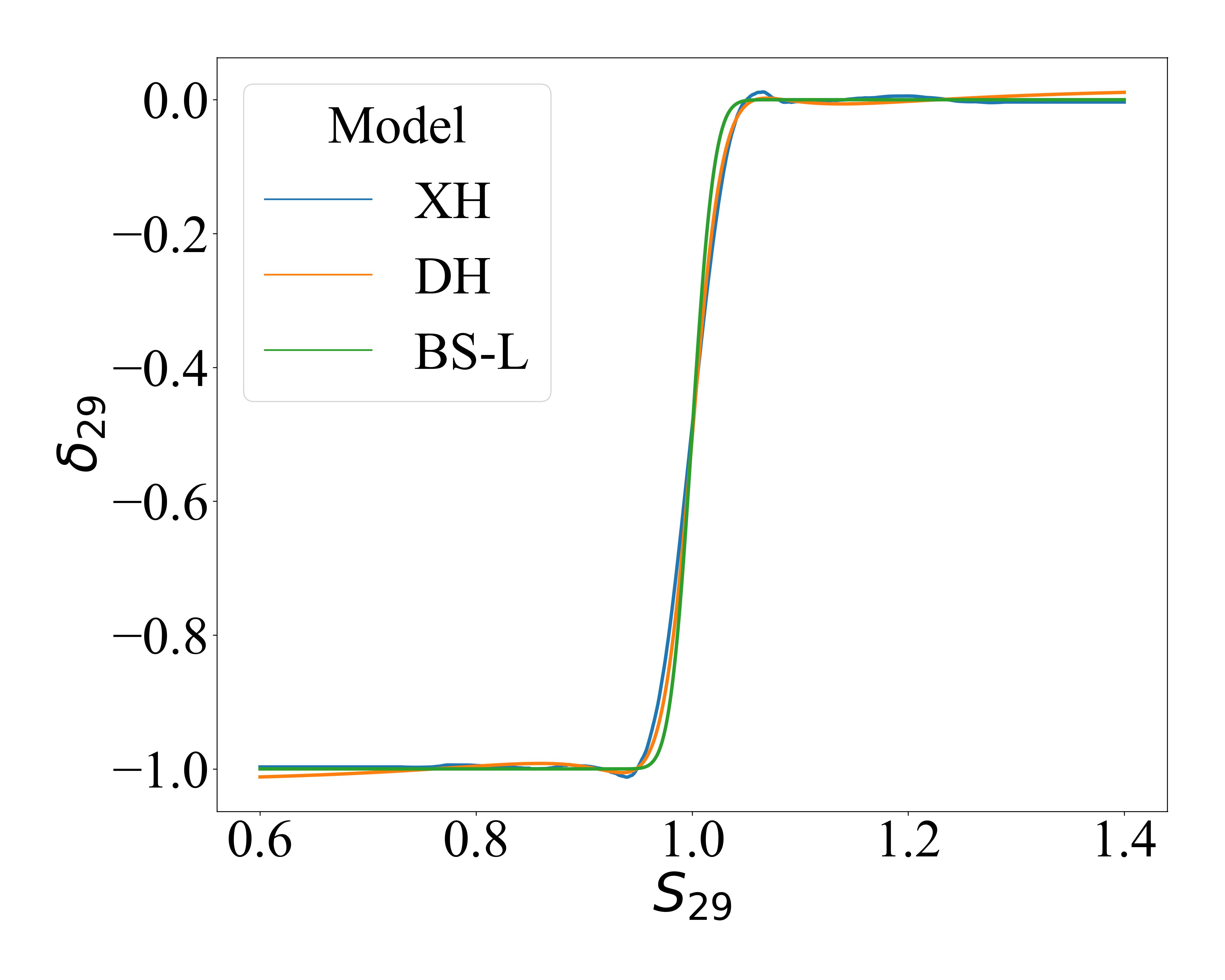}
\caption{Hedging strategies for ATM put option at times $k=1,10,15,20,25,29$ under transaction costs and the CTM$_2$ loss function.
\label{fig:strategy-proportional-cvar-square-put}}
\end{minipage}
\end{center}
\end{figure}

We present the results from experiments for X Hedging, Deep Hedging, and BS-L Hedging in terms of the statistical properties (Table \ref{tab:results-call-put-atm}), the histograms of the $\text{P\&L}_T$ distributions (Figure \ref{fig:histogram-proportional-cvar-square}),  and the plots of hedging strategies at various time epochs (Figures \ref{fig:strategy-proportional-cvar-square-call} and  \ref{fig:strategy-proportional-cvar-square-put}).

In Table \ref{tab:results-call-put-atm}, for both call and put options, all hedging models exhibit standard deviation of similar magnitude and mean close to the target value of zero, indicating that they are effectively producing optimal hedging strategies. 
The JS-divergence value between Deep Hedging and X Hedging is as small as $0.000146$ (call) and $0.000111$ (put), indicating a high degree of similarity in the $\text{P\&L}_T$ distributions. In addition, X Hedging demonstrates significantly faster computation times, further supporting the advantages of the proposed framework in this setup. 

\begin{table}[h]
\caption{Terminal $\text{P\&L}_T$ summary statistics for ATM options under CTM$_2$ loss.}
\label{tab:results-call-put-atm}
\begin{tabular}{@{}llrrrrr@{}}
\toprule
 & & \textbf{Mean} & \textbf{Std} & \textbf{JS(DH$\|\cdot$)} & \textbf{JS(XH$\|\cdot$)} & \textbf{Time (sec)} \\
\midrule
 & BS-L & -0.002238 & 0.005648 & 0.004041 & 0.004754 & 614.0 \\
\textbf{Call} & DH & -0.002007 & 0.005888 & --- & 0.000146 & 42616.0 \\
 & XH & -0.002164 & 0.005845 & --- & --- & 38887.0 \\
\midrule
 & BS-L & -0.002667 & 0.004229 & 0.002015 & 0.004007 & 889.0 \\
\textbf{Put} & DH & -0.003753 & 0.006070 & --- & 0.000111 & 45887.0 \\
 & XH & -0.002260 & 0.004599 & --- & --- & 32346.0 \\
\botrule
\end{tabular}
\footnotetext{The table reports the mean, standard deviation (Std), and Jensen--Shannon (JS) divergence of terminal P\&L distributions under transaction costs.}
\end{table}

Comparing the $\text{P\&L}_T$ distribution histograms for the three hedging models in Figure \ref{fig:histogram-proportional-cvar-square}, we observe that, for both call and put options, the distributions resemble a normal distribution with a slightly negative mean and a longer left tail.

The hedging strategies at several selected time epochs, namely $t_k$, for $k=1, 10, 15, 20, 25, 29$ are shown in Figures \ref{fig:strategy-proportional-cvar-square-call} and \ref{fig:strategy-proportional-cvar-square-put} for the call and put options, respectively. Here, $S_k$ denotes the underlying asset value, and $\delta_k$ represents the corresponding hedging strategy at time index $k$. 
Since both figures display consistent patterns and convey similar insights, we focus our detailed analysis on Figure \ref{fig:strategy-proportional-cvar-square-call}. 

Note that the cases $k = 0$ and $k = 30$ are not informative to visualize: at $k = 0$, all simulated paths share the same $S_0$, resulting in a single-point strategy; at $k = 30$, no hedging occurs, as the terminal $\text{P\&L}_T$ is computed at this final time step.
At intermediate time epochs, specifically those shown here, $t_k$, for $k=10, 15, 20, 25$, the hedging strategies from X Hedging and Deep Hedging closely approximate the delta strategies from BS-L Hedging. 
At $k = 1$, X Hedging and Deep Hedging differ most from BS-L Hedging, particularly for small and large values of the underlying asset price. This is expected since the simulated data does not reach these ranges for small $k$ values. 
X Hedging is almost flat in these intervals, corresponding to the leaves with the lowest and highest hedging strategies. However, the seemingly inaccurate strategies at early time steps do not significantly affect the final $\text{P\&L}_T$ because very few paths reach the low and high values. 
Indeed, when $k$ approaches $29$, they take the form of step functions.  
This is natural, as predicting where each path will end up is more challenging at earlier time steps than those closer to maturity. Consequently, during the early time steps, few decisive hedging actions are taken, whereas in the later stages, the hedging models tend to either hedge extensively or not at all.

In general, X Hedging strategies tend to converge towards the BS-L Hedging as $t$ approaches maturity. Although there are differences at the smallest and largest values of $S_k$, X Hedging generally produces accurate strategies. 
To assess the robustness of the model's performance, we conduct similar experiments for in-the-money (ITM) and out-of-the-money (OTM) call and put options. These tests yield consistent insights and conclusions, and we summarize the results in Appendix \ref{app:allmoneyness}. 
Across the preceding experiments, we find that X Hedging performs comparably to both Deep Hedging and Black-Scholes benchmarks. Thus, we see no indication that Deep Hedging is preferable to X Hedging in the setups we considered. This establishes a solid foundation for X Hedging's performance, making it even more relevant to address the user-centered explainability issue.

\section{Comparative advantages of X Hedging}\label{Sec:CompaAdv}

While Deep Hedging has attracted considerable attention due to its flexibility and performance in high-dimensional settings, X Hedging offers some distinct advantages that make it a valuable alternative, particularly in practical and data-constrained environments. In this section, we highlight two key benefits of X Hedging in comparison to Deep Hedging, namely its ability to perform well with limited data and its enhanced user-centered explainability.

\subsection{Resilience to data constraints}

As noted by the original authors of Deep Hedging in \citet{deephedging_vae}, one limitation is its reliance on a large training dataset to learn effective hedging strategies because neural networks have many degrees of freedom.
X Hedging, which is based on tree ensemble methods, is designed to be more data-efficient. Unlike Deep Hedging, it can be trained on relatively small datasets while still achieving robust performance. 
%

Across all our experiments, X Hedging consistently demonstrates significantly greater robustness to limited data availability. Specifically, it maintains strong performance even when the number of training sample paths is reduced by a factor of ten. In contrast, neural network-based Deep Hedging shows a marked decline in performance under the same conditions. This advantage in small-data settings is observed consistently throughout the study.

\begin{figure}[h]
\centering

\begin{minipage}[t]{0.32\textwidth}
\centering
\includegraphics[width=\linewidth]{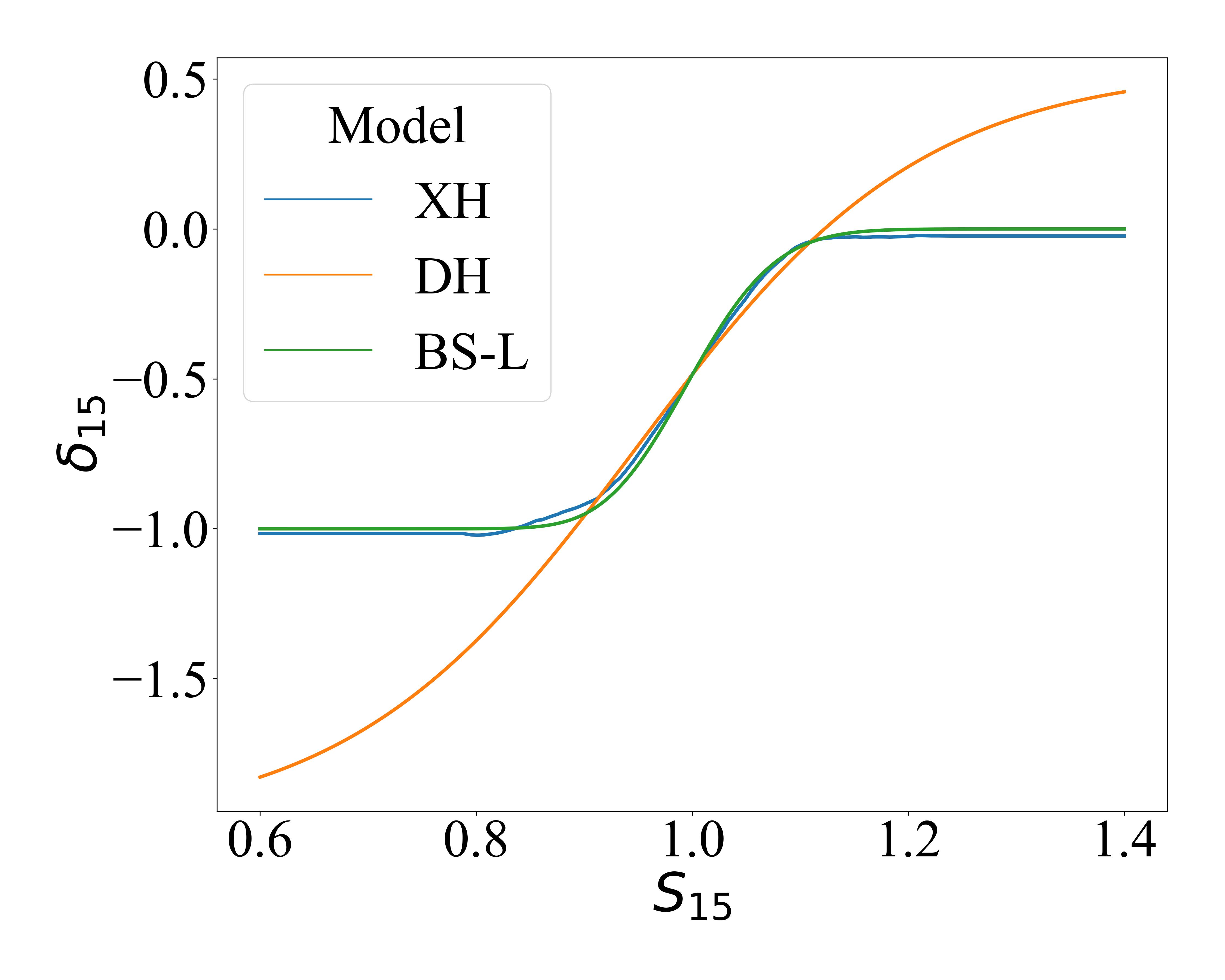}\\
(a) $k=15$.
\end{minipage}
\hfill
\begin{minipage}[t]{0.32\textwidth}
\centering
\includegraphics[width=\linewidth]{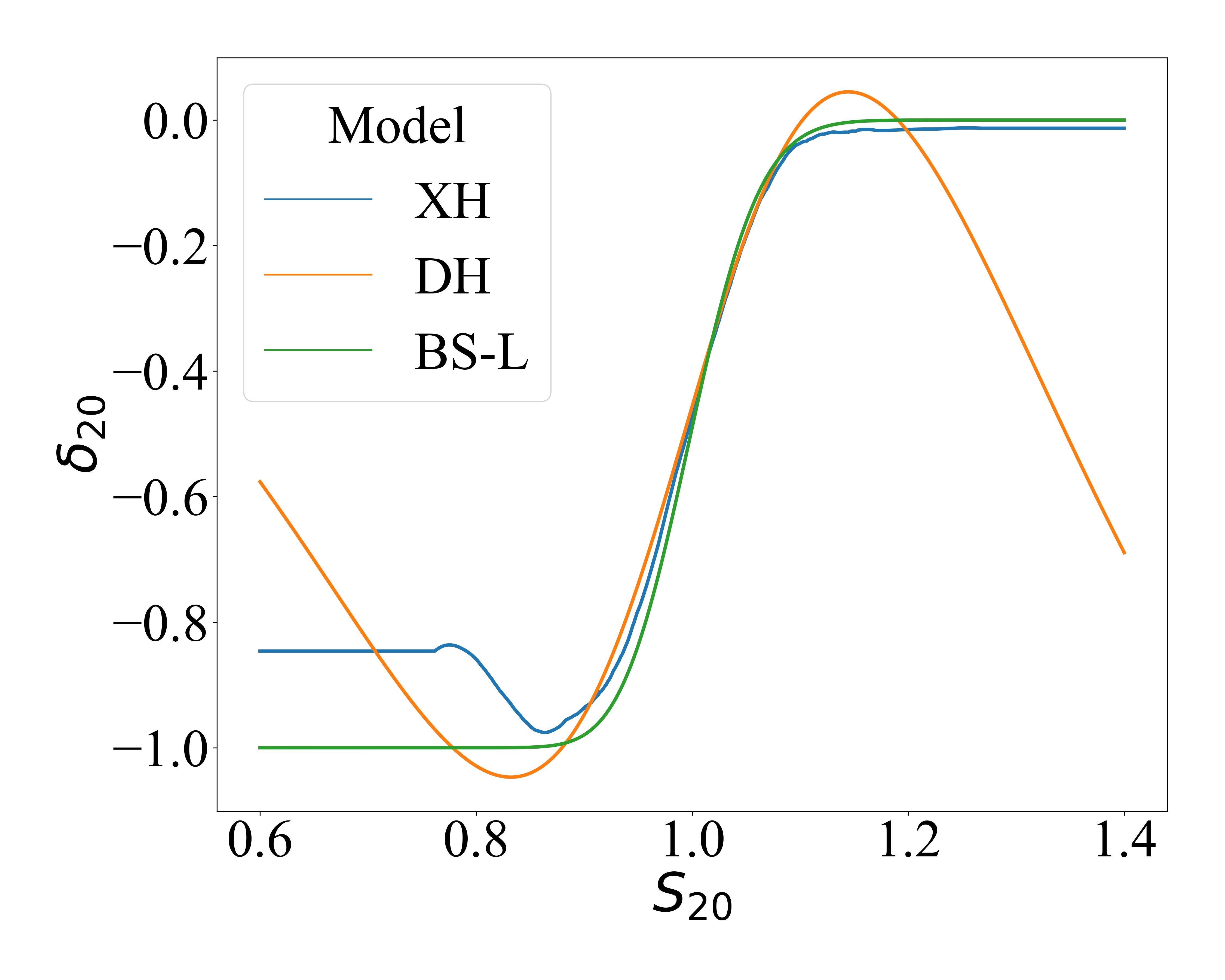}\\
(b) $k=20$.
\end{minipage}
\hfill
\begin{minipage}[t]{0.32\textwidth}
\centering
\includegraphics[width=\linewidth]{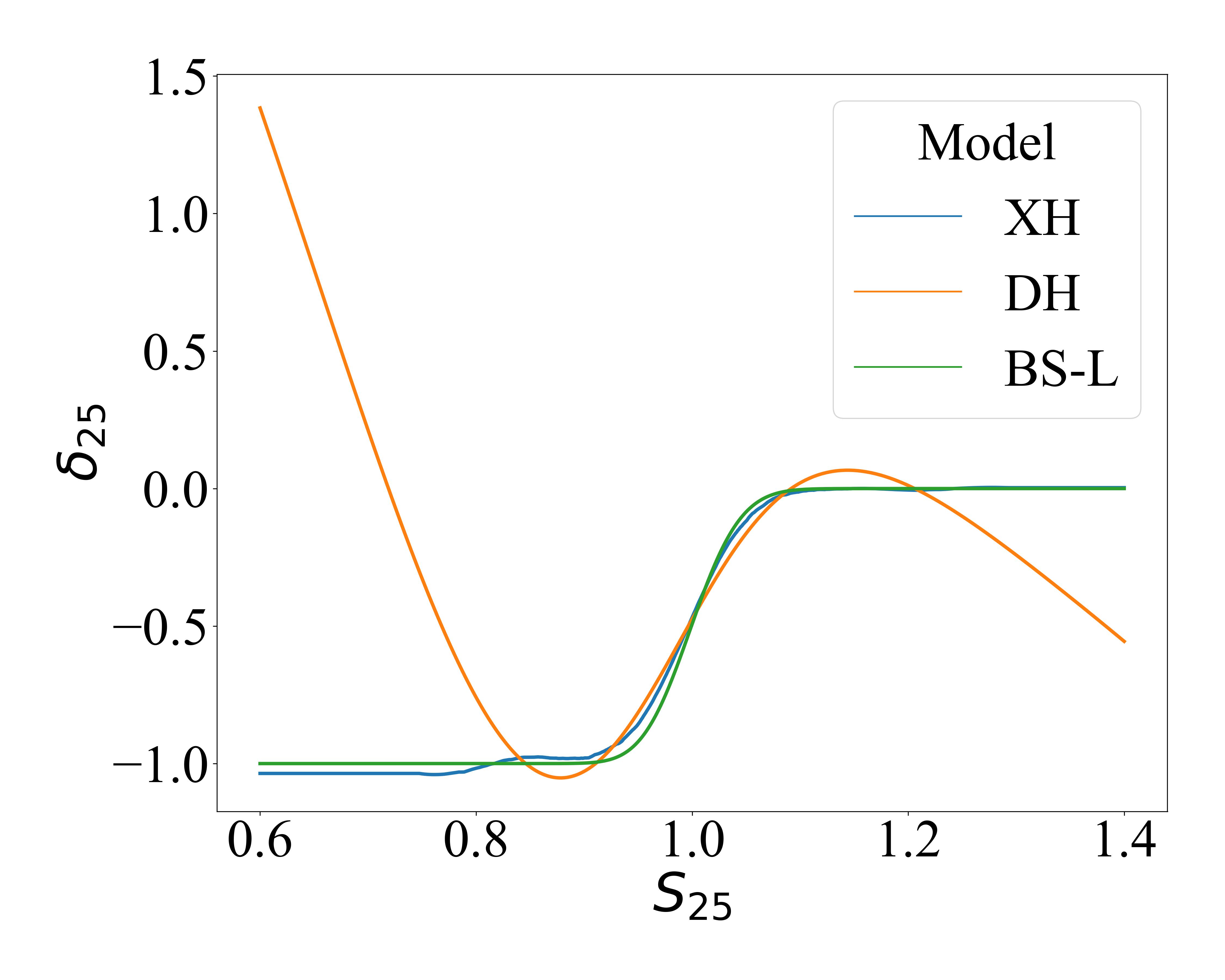}\\
(c) $k=25$.
\end{minipage}

\caption{Hedging strategies for an ATM put option with data constraints at times $k=15,20,25$ under proportional transaction costs and the CTM$_2$ loss function.}
\label{fig:strategy-proportional-cvar-square-put-20000}
\end{figure}

We demonstrate it using an example of an ATM put option under proportional transaction costs and using the CTM$_2$ loss function.  Under the same setup as in Section \ref{sec:ValidationCVaR}, we reduce the number of training sample paths from $N = 200 000$ to $N=20 000$ and compare the performance of Deep Hedging and X Hedging. 
Figure \ref{fig:strategy-proportional-cvar-square-put-20000} illustrates this behavior for three selected time epochs. 
Deep Hedging often deviates substantially from the benchmark hedging strategies, particularly in the deep in-the-money and deep out-of-the-money regions, and sometimes in both.  In contrast, X Hedging consistently approximates hedging strategies with high accuracy even with this smaller number of training sample paths. 

\subsection{User-centered explainability} 

X Hedging is based on a tree ensemble model that, although it does not provide the full explainability of a single decision tree, still achieves a high degree of user-centered explainability, as shown by \citet{herm2023stop}. This is a clear advantage over Deep Hedging, which relies on neural networks and is typically considered a black-box method. It lacks explainability across all dimensions unless complemented by an external XAI methodology.

\begin{figure}
\begin{center}
\begin{minipage}{\textwidth}
\centering
\includegraphics[width=\textwidth]{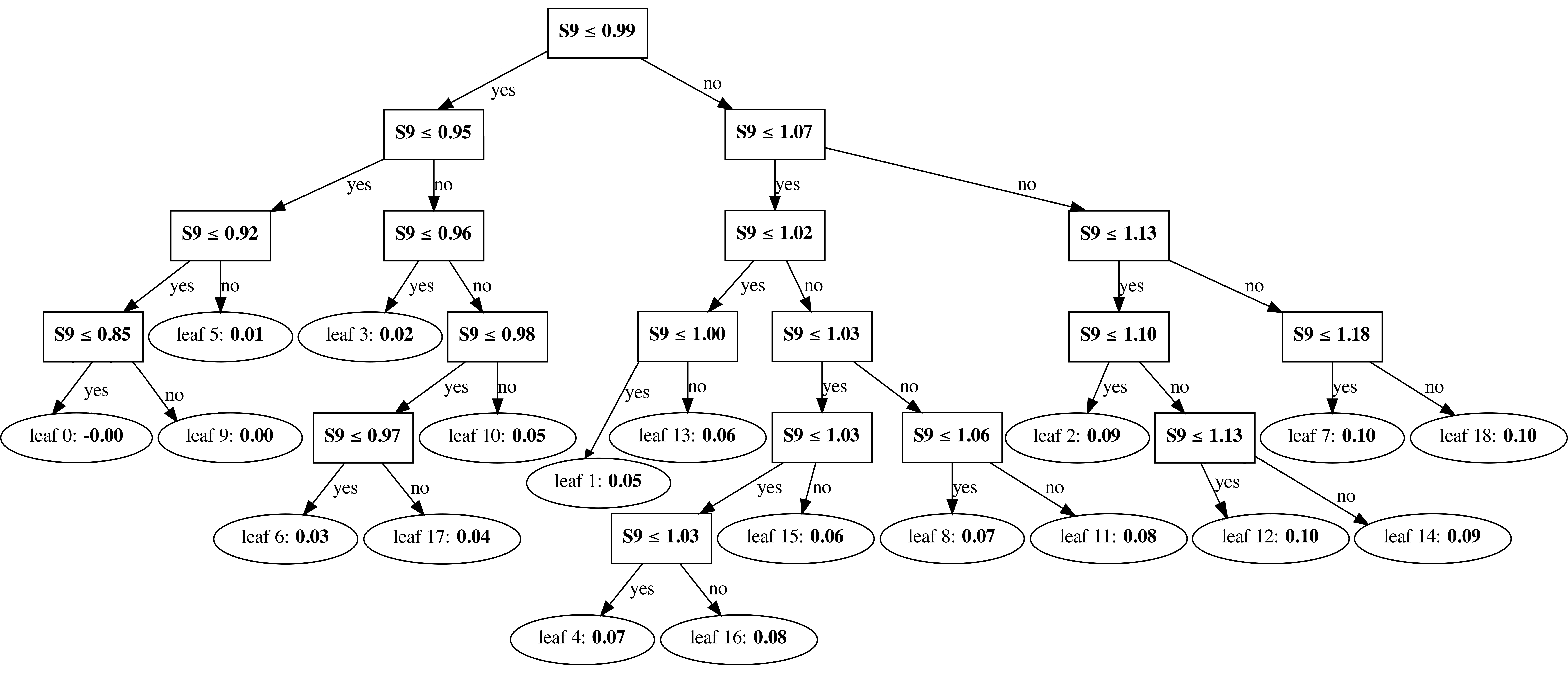}
\caption{One of the X Hedging decision trees visualized at time 9.
\label{fig:tree_s9}}
\end{minipage}
\end{center}
\end{figure}

To illustrate how X Hedging achieves user-centered explainability, we now describe the structure and decision logic of the underlying tree model. For simplicity, we set $n=10$ and $N=20000$ in the illustrations presented in this section. In X Hedging, each node in the tree contains a condition on one of the input features, and the child node depends on whether the condition is fulfilled or not. This means that every node relates to a condition on $S_k$, and the proceeding nodes depend on whether the condition is met by the value of $S_k$. This structure makes it easy to explain a hedging strategy at time step $t_k$. The hedging strategy is obtained by traversing all trees in $G_k$ until a leaf node is reached. Ultimately, the model's final decision is a linear combination of multiple trees in the ensemble, but each decision can still be backtracked. An example of one of the fitted trees is shown in Figure \ref{fig:tree_s9}, which highlights the explainability  
offered by the model.
The intermediate decisions from input to output are easy to follow, allowing us to understand how changes in the input affect the outcome. This level of insight enhances trust and confidence in the hedging model and increases its likelihood of meeting regulatory and institutional demands for explainability.

\begin{figure}
\begin{center}
\begin{minipage}{\textwidth}
\centering
\includegraphics[width=0.9\textwidth]{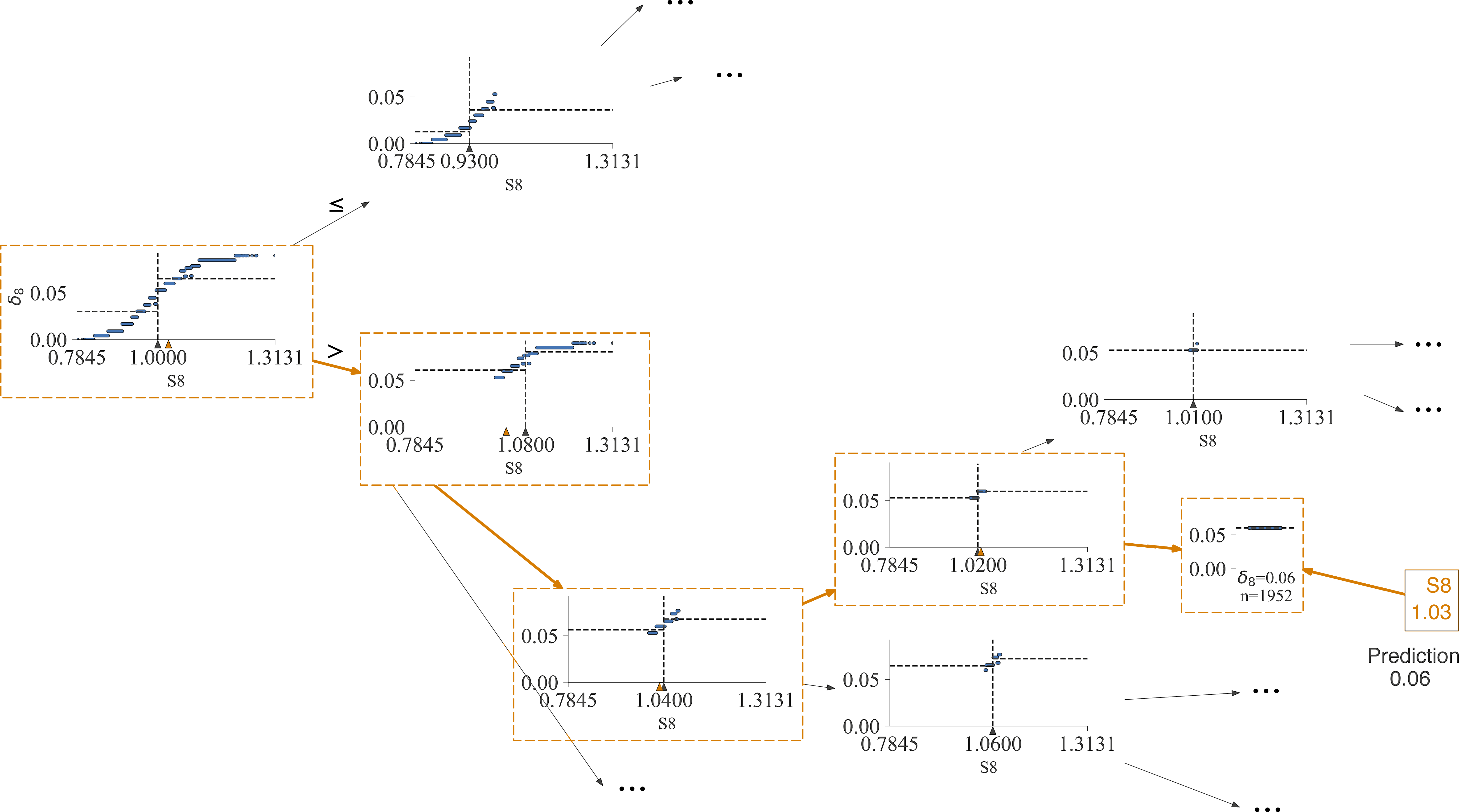}
\caption{Prediction path in one tree for $S_8$ equal to 1.03.
\label{fig:tree-path}}
\end{minipage}
\end{center}
\end{figure}

In Figure \ref{fig:tree-path}, 
we visualize how decisions are made within a section of one of the trees in X Hedging at time $t_8$, along with information about how the hedging strategies are distributed along the tree.  
The hedging strategy curve is split throughout the tree, and each data point in the curve ends up in a leaf node, where it is associated with a specific hedging strategy, $\delta_8$. The data points are the underlying asset path values. It is worth noting that the hedging strategy curve in the tree's root strongly resembles the hedging strategies presented in Section \ref{sec:ValidationCVaR}. Figure \ref{fig:tree-path} also shows how many strategies belong to each leaf node, which is useful information to characterize common prediction paths throughout the tree. The leaves in this example contain between $21$ and $1952$ data points. The leaves with the most data points are around $S_8 = 1.0$, reflecting the zero interest rate. Leaves with few data points can be treated as outliers, and hedgers should pay attention to them. A smaller change to $S_8$ might change which leaf the data point ends in. Hence, tree-based models offer a way to explain output certainty. Leaves with more training data offer more reliable predictions than leaves with fewer samples. 

To demonstrate the user-centered explainability of X Hedging, we have highlighted a prediction path for $S_8 = 1.03$ with orange boxes and arrows in Figure \ref{fig:tree-path}. This shows all decisions the tree makes to predict $\delta_8=0.06$. It is easy to follow all decisions since they only compare numerical values. Since strategy curves are plotted with the vertical decision line, it is also possible to see how close the data point is to ending up in another leaf. This is, for example, the case in the third node of the highlighted prediction path. A small increase of $S_8$ will make the data point end up in the other successive node. As this example shows, trees offer an intuitive way to explain the hedging model's decisions.

\section{X Hedging under Jump-Diffusion Dynamics}
\label{xhedging_with_jumps}
Having established the comparative performance of X Hedging and Deep
Hedging across the experiments in
Sections~\ref{sec:performance_comparison} and~\ref{Sec:CompaAdv}, we now
examine the robustness of X Hedging under more challenging market
dynamics. Specifically, we consider the Merton jump-diffusion model
introduced in Equation~\eqref{jump_model} in Section~\ref{sec:methodology},
which allows for discontinuous asset-price movements and provides a more
realistic description of extreme price changes. The purpose of this section is not to provide an additional comparison with Deep Hedging, but to assess the robustness and practical reliability of X Hedging under more turbulent market conditions using a model-based benchmark hedging strategy.

To analyze the impact of jumps on X Hedging performance, we consider a 30-day at-the-money call option with daily rebalancing. To isolate the jump effect, we assume a frictionless market with no transaction costs. The parameters used to simulate the underlying asset under the jump diffusion model in Equation (\ref{jump_model}) are summarized in Table \ref{tab:parametersJump}. The benchmark hedging model (BH) is given by the semi-analytical formula in Equation (\ref{delta_jump}).

\begin{table}[h]
\caption{Jump-diffusion model parameters for asset price simulation and option parameters.}
\label{tab:parametersJump}
\begin{tabular}{@{}ccccccccccc@{}}
\toprule
$S_0$ & $K$ & $r$ & $d$ & $\sigma$ & $\lambda_J$ & $\mu_J$ & $\sigma_J$ & $T$ & $n$ & $N$ \\
\midrule
$1$ & $1$ & $0.0$ & $0.0$ & $0.3$ & $0.5$ & $-0.1$ & $0.2$ & $\frac{30}{365}$ & $30$ & $200000$ \\
\botrule
\end{tabular}
\end{table}

\begin{figure}
\begin{center}
\begin{minipage}{\textwidth}
\centering
\includegraphics[width=0.70\textwidth]{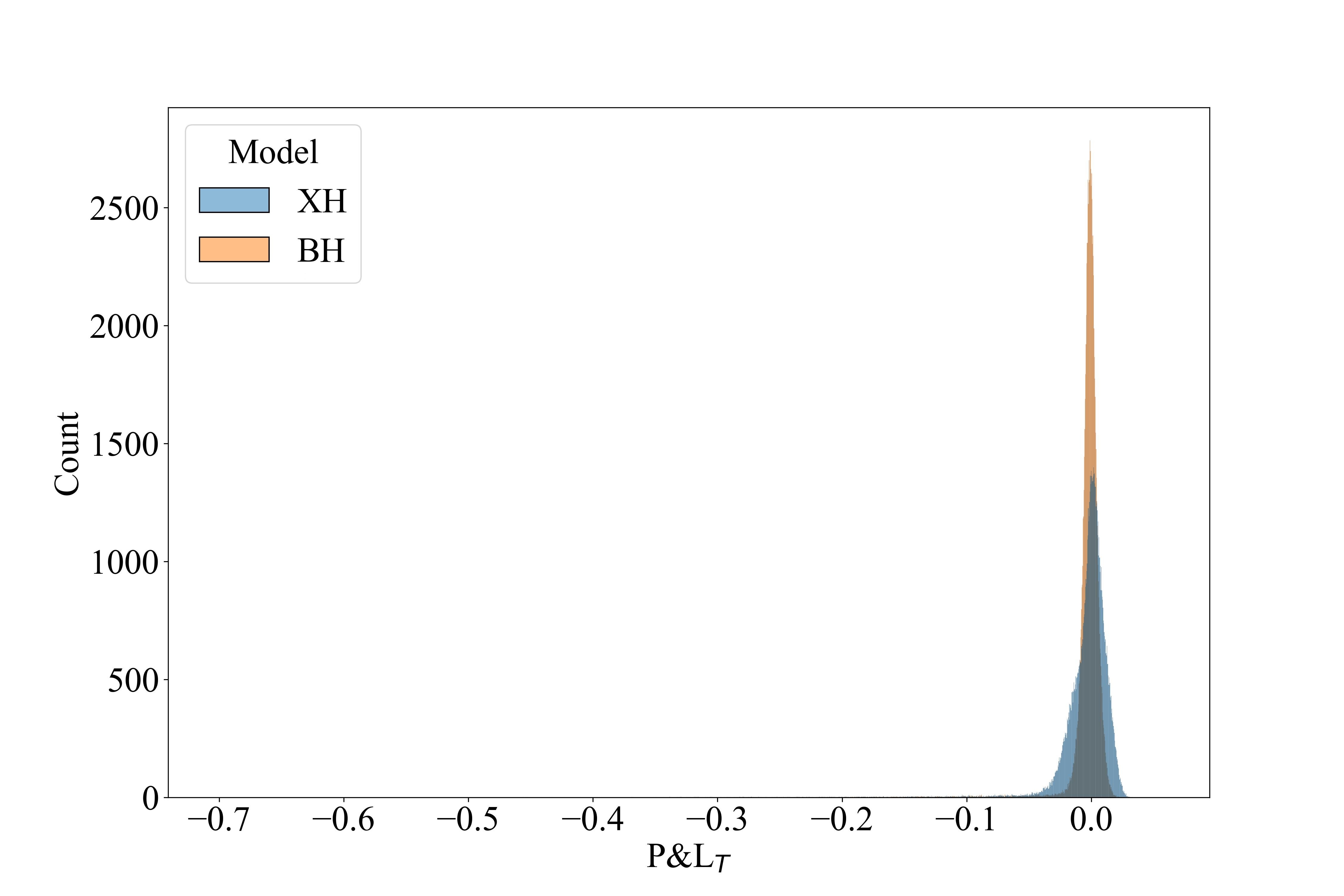}
\caption{Histograms of the P\&L$_T$ distributions for benchmark (BH) and X Hedging (XH).
\label{fig:histJump}}
\end{minipage}
\end{center}
\end{figure}

Figure \ref{fig:histJump} shows the histograms of the $\text{P\&L}_T$ distributions for both the benchmark model and the X Hedging strategy under the jump-diffusion framework, while Table \ref{tab:DescStatsJump} reports the corresponding descriptive statistics.

\begin{table}[h]
\caption{Summary statistics of the $\text{P\&L}_T$ for the benchmark and the X Hedging strategies in a jump-diffusion setting.}
\label{tab:DescStatsJump}
\begin{tabular}{@{}lrrrrrr@{}}
\toprule
\textbf{Model} & \textbf{Mean} & \textbf{Std Dev} & \textbf{Skewness} & \textbf{Kurtosis} & \textbf{Min} & \textbf{Max} \\
\midrule
BH & -0.003102 & 0.018910 & -10.830045 & 173.837940 & -0.703028 & 0.023749 \\
XH & -0.002862 & 0.019757 & -6.600901 & 90.950098 & -0.681716 & 0.056935 \\
\botrule
\end{tabular}
\footnotetext{BH denotes the benchmark hedging strategy; XH denotes X Hedging.}
\end{table}

The histograms and descriptive statistics clearly indicate that both hedging models exhibit significantly more extreme losses when jumps are present. This deterioration in hedging performance when jump dynamics are introduced is not unexpected and can be attributed to several factors inherent in the design of the current hedging setups. 
Delta hedging, by its nature, assumes continuous and small price changes and therefore fails to neutralize the risk associated with sudden, discontinuous movements in the underlying, a characteristic feature of jump-diffusion processes. When a jump occurs, the portfolio cannot be rebalanced instantaneously, leading to hedging errors that are amplified precisely when market conditions are most adverse. 
Incorporating sensitivity to higher-order Greeks, such as gamma- and vega-hedging, could potentially improve hedging effectiveness in jump-diffusion settings by accounting for convexity and volatility exposure, which delta alone cannot capture. 
%


Additionally, both hedging models perform delta hedging using only the underlying asset. This limitation of having a single hedging instrument further constrains the model’s ability to manage risk under jump scenarios. In practice, effective hedging of jump risk typically requires access to options or other nonlinear instruments that are sensitive to tail events. Without such tools, the models lack the means to replicate or offset the potential impact of rare but significant price discontinuities. This structural limitation highlights an important boundary of the current framework and explains, at least in part, the increased hedging error observed under jump-augmented simulations.

Although both hedging models experience extreme losses, both yield small average losses (mean $\text{P\&L}_T$ around -0.3\%) with comparable standard deviations. However, important differences in the tails merit attention. Under the pure diffusion scenario (GBM), the  $\text{P\&L}_T$ histograms for both strategies exhibit tight clustering around zero, with relatively thin and symmetric tails. This suggests hedging errors are well contained and that both methods perform comparably in a continuous-path environment. However, introducing jumps, particularly with a negative mean jump of -0.1, substantially alters the distributional characteristics. The histograms become sharply left-skewed with noticeably heavier tails, reflecting a higher frequency and severity of significant losses. The benchmark strategy, in particular, shows a distinct peak near zero combined with a heavy left tail, underscoring its susceptibility to extremely negative  $\text{P\&L}_T$ outcomes. While the X Hedging model is also affected by jumps, it shows a more resilient profile, with a less pronounced tail and fewer extreme losses. This is supported by a higher minimum  $\text{P\&L}_T$ value and more favorable skewness and kurtosis statistics.
To further compare the performance of both methods in the tails, we compute the VaR and the CVaR at various confidence levels. The results are presented in Table \ref{tab:CVar}.

\begin{table}[h]
\caption{VaR and CVaR for the benchmark and the X Hedging assuming different confidence levels.}
\label{tab:CVar}
\begin{tabular}{@{}ccccc@{}}
\toprule
\textbf{Confidence level} & \multicolumn{2}{c}{\textbf{VaR}} & \multicolumn{2}{c}{\textbf{CVaR}} \\
(\%) & BH & XH & BH & XH \\
\midrule
99.5 & -0.136173 & -0.118569 & -0.207796 & -0.182217 \\
99.0 & -0.083993 & -0.075401 & -0.157489 & -0.138550 \\
97.5 & -0.020942 & -0.036190 & -0.089398 & -0.084914 \\
\botrule
\end{tabular}
\footnotetext{BH denotes the benchmark hedging strategy; XH denotes X Hedging.}
\end{table}

The previous results suggest that X Hedging delivers more stable performance in the presence of rare but severe shocks compared to the benchmark strategy. The downside risk measures also support this robustness. At the 99.5\% confidence level, the CVaR for the benchmark reaches $-20.78\%$, whereas X Hedging contains losses to $-18.22\%$, a difference of 256 basis points. At the 99\% level, the gap remains substantial, with CVaR values of $-15.75\%$ for the benchmark and $-13.86\%$ for X Hedging, representing a difference of 189 basis points. 
These results suggest that X Hedging can offer a modest advantage in managing downside risk under extreme market conditions characterized by jumps. While the benchmark strategy may appear slightly more efficient in normal environments, it tends to exhibit larger losses in the tails. X Hedging shows somewhat greater resilience in these scenarios. However, it is not immune to significant losses either. As discussed earlier, the model’s limitations, particularly the use of a single hedging instrument and the exclusive reliance on delta hedging, restrict its ability to fully mitigate risks arising from discontinuous price movements. Therefore, while X Hedging performs comparatively better in the presence of jumps, it should not be viewed as an ideal solution but rather as a step toward more robust approaches to hedging under market discontinuities.

\section{Conclusion} 
\label{sec:conclusion}

This paper introduces X Hedging, a novel alternative to the Deep Hedging framework that addresses the critical explainability gap in contemporary AI-driven hedging approaches. By replacing neural networks with more explainable tree-ensemble methods, specifically LightGBM, X Hedging achieves performance comparable to Deep Hedging while taking a step toward more explainable, user-centered AI-based hedging. 
This feature is particularly important in meeting the growing regulatory requirements for transparency and accountability in the financial sector. 

We establish that LightGBM, a model that has demonstrated strong performance in other domains, is also a viable and effective choice for hedging. To leverage LightGBM in this setting, our framework incorporates a novel optimization technique for integrating tree ensemble models into a reinforcement learning setup, overcoming methodological challenges such as the absence of stochastic gradient descent and backpropagation. The results demonstrate that X Hedging effectively balances performance and explainability, making it a practical alternative for institutions that must deploy hedging models under governance and compliance constraints.

By adapting tree ensemble methods to a Deep Hedging–inspired framework, this work highlights their potential to narrow the gap between state-of-the-art AI methods and the practical requirements of financial decision-making. 
In addition to achieving hedging performance comparable to Deep Hedging, X Hedging offers significant advantages in computational efficiency and its ability to learn effectively from limited data, properties that further support its suitability for real-world deployment beyond explainability alone. 

X Hedging thus represents an AI-based approach to hedging that moves the frontier toward more explainable and governable models, while retaining the benefits of data-driven learning. 
Several avenues could deepen the realism and applicability of the X Hedging framework. These include extending the approach to other financial instruments, incorporating alternative risk measures, or further improving the optimization methodology to enhance scalability and efficiency.  Another important direction is including bonds as hedging instruments to account for non-zero interest rate environments, allowing the model to better reflect real-world financial conditions. Additionally, adopting more sophisticated models for the underlying asset's dynamics, such as stochastic volatility models with jumps or multifactor stochastic volatility models, would enable the framework to capture more complex market behavior. 
A promising line of research is to strengthen the model’s ability to handle jump risk more effectively, either by including additional hedging instruments, such as options, to better manage tail exposure or by modifying the learning objective to place greater emphasis on rare events.  
Such extensions would not only enhance performance under extreme scenarios but also align the framework more closely with practical risk management requirements in volatile market environments. Finally, applying X Hedging in empirical hedging settings using real market data represents an important step toward assessing its robustness and operational relevance. 

\section*{Declarations}

\subsection*{Funding}
Rita Pimentel was partially funded by the COMPAMA project from The Research Council of Norway, grant number 314609.

\subsection*{Conflict of interest}
The authors report no conflicts of interest. The authors alone are responsible for the content and the writing of the paper.

\subsection*{Ethics approval and consent to participate}
Not applicable.

\subsection*{Consent for publication}
Not applicable.

\subsection*{Data availability}
No external datasets were used in this study. The numerical data were generated through simulations using the models and methods described in the manuscript.

\subsection*{Materials availability}
Not applicable.

\subsection*{Code availability}
A GitHub repository for the X Hedging framework is available at
\href{https://github.com/HaidaQuant/Xhedging}{GitHub}. The complete implementation will be made publicly
available upon acceptance of the manuscript.

\subsection*{Author contribution}
This work is based on the master's thesis \citet{key} by B.A.A. and J.B., supervised by Y.N. and R.P. B.A.A. and J.B. conducted all original implementations and drafted the manuscript, with methodology and results interpretation discussed in weekly meetings with Y.N. and R.P., who also provided feedback on the writing process. Subsequently, A.H. took primary responsibility for reviewing and improving the existing implementation and conducting extensive new numerical experiments, which provided additional results and led to new findings reported in the manuscript. Y.N. and R.P supervised the additional computational work and improved and reshaped the thesis into a manuscript for submission.  

\begin{appendices}

\clearpage
\section{X Hedging pseudocode}
\label{pseudocode}

\setcounter{algorithm}{0}
\renewcommand{\thealgorithm}{A.\arabic{algorithm}}

\begin{algorithm}[h]
\caption{X Hedging}
\label{alg:dhlightgbm}
\begin{algorithmic}[1]

\Procedure{XHedging}{$S, K, \kappa, \ell, J$}

\State $n$: time steps; $N$: paths; $K$: strike
\State $S \in \mathbb{R}^{n\times N}$: simulated prices
\State $Z \gets \max \{S_T-K,0\}$ or $\max \{K-S_T,0\}$
\State $\kappa$: transaction cost constant; $\ell$: loss (MSE or CTM$_2$)
\State $G_1,\dots,G_{n-1}$: LightGBM models

\Statex
\For{$k \gets n-1$ \textbf{down to} $1$}
    \State $\delta_k \gets G_k(S_k)$
    \State compute $\text{P}\&\text{L}_T$, $\ell$, gradient, Hessian; update $G_k$
\EndFor

\Statex
\State $R \gets$ random permutation of $\{1,\dots,n-1\}$
\For{$j \gets 1$ \textbf{to} $J$}
    \For{$k$ \textbf{in} $R$}
        \State $\delta_k \gets G_k(S_k)$
        \State compute $\text{P}\&\text{L}_T$, $\ell$, gradient, Hessian; update $G_k$
    \EndFor
\EndFor

\State \textbf{return} $\{G_k\}_{k=1}^{n-1}$, $\text{P}\&\text{L}_T$

\EndProcedure
\end{algorithmic}
\end{algorithm}

\clearpage
\section{Parameters used for the hedging frameworks}
\label{app:parameters}

The parameters for X Hedging are presented in Table \ref{tab:parameters-lgbm}. 
The minimum number of data points in each leaf node is given by \textit{N. data leaf}. Larger values will result in lower variance. The maximum number of leaves for each tree, where $j$ is the current iteration, is given by \textit{N. leaves}. At each iteration, the maximum number of leaves increases to create a more complex model without overfitting early on. The number of boosting rounds used to reduce the loss of the model output, where $j$ is the current iteration, is given by \textit{N. boost rounds}. The number of boosting rounds increases for every iteration to create a more complex model. The learning rate multiplier for each weak learner is the \textit{learning rate}. The maximum number of boosting rounds without loss improvement is set by \textit{early stopping}. The number of times we randomly iterate over each model at each time step is given by \textit{iterations}.

\begin{table}[h]
\caption{Parameters used for the LightGBM models in X Hedging.}
\label{tab:parameters-lgbm}
\begin{tabular*}{\textwidth}{@{\extracolsep\fill}cccccc}
\toprule
\textbf{N. data leaf} &
\textbf{N. leaves} &
\textbf{N. boost rounds} &
\textbf{Learning rate} &
\textbf{Early stopping} &
\textbf{Iterations} \\
\midrule
$5$ & $10+j$ & $10+5j$ & $0.1$ & $5$ & $15$ \\
\botrule
\end{tabular*}
\end{table}

The parameters for Deep Hedging are presented in Table \ref{tab:parameters-nn}. The \textit{activation function} tanh is used due to its similarities with the Black--Scholes hedging strategies, which should make finding the correct strategies more efficient. The stochastic gradient descent \textit{optimizer} Adam is chosen due to its well-documented performance. The initial learning rate of $0.01$ is chosen as a standard value. The size of the neural networks, the number of \textit{neurons}, the \textit{batch size}, and the number of \textit{epochs} were defined based on the accuracy provided in our experiments.

\begin{table}[h]
\caption{Parameters used for the neural networks in Deep Hedging.}
\label{tab:parameters-nn}
\begin{tabular*}{\textwidth}{@{\extracolsep\fill}cccccc}
\toprule
\textbf{Activation function} &
\textbf{Optimizer} &
\textbf{Initial learning rate} &
\textbf{Neurons} &
\textbf{Batch size} &
\textbf{Epochs} \\
\midrule
tanh & Adam & $0.01$ & $32$ & $1024$ & $10000$ \\
\botrule
\end{tabular*}
\end{table}

\clearpage
\section{Results for hedging strategies with no transaction costs and MSE loss function}
\label{app:baseCase1}
\setcounter{figure}{0}
\setcounter{table}{0}

\begin{figure}[h]
\centering
\includegraphics[width=0.8\textwidth]{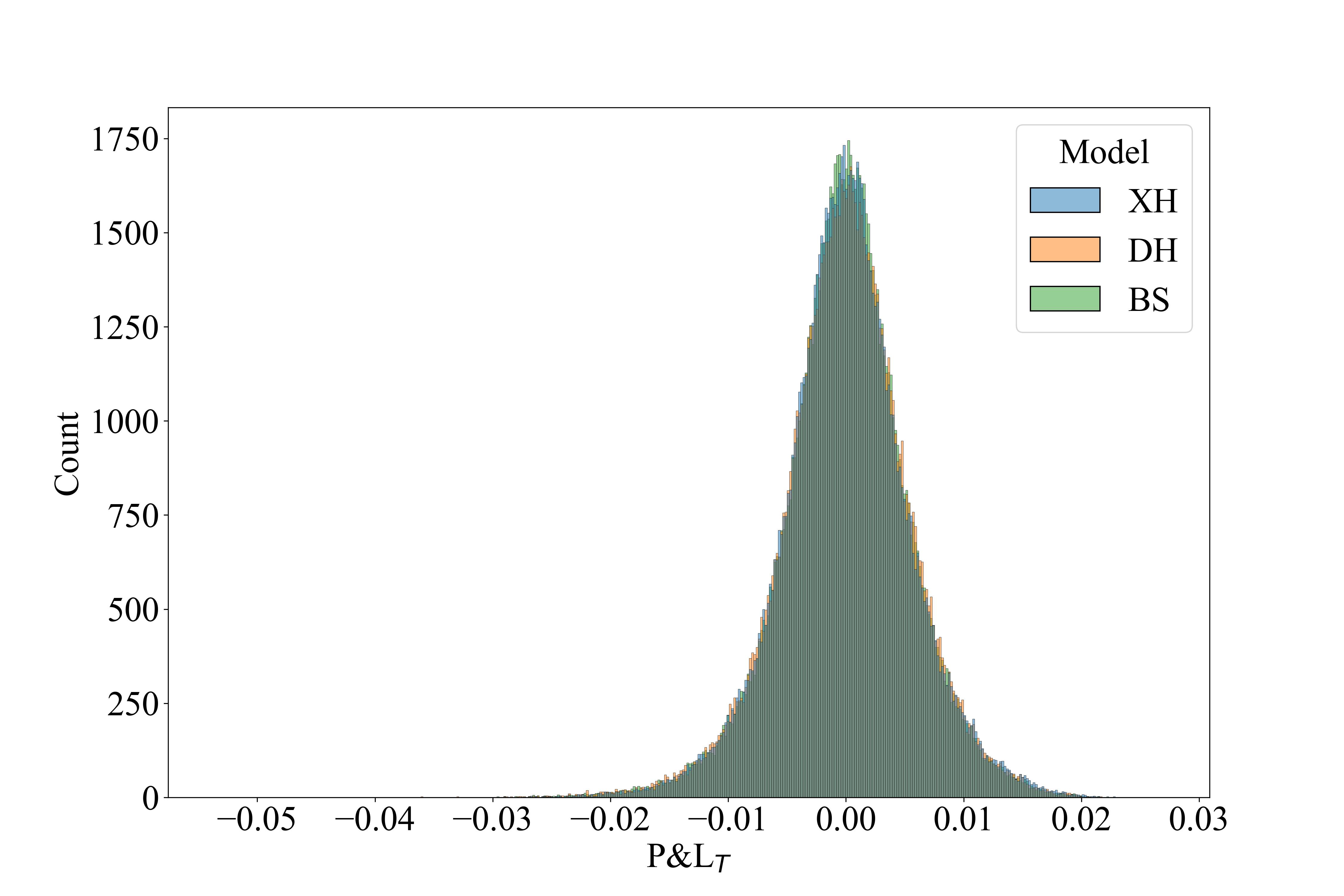}
\caption{Histograms of the $\text{P\&L}_T$ distributions without transaction costs and using the MSE loss function.}
\label{fig:histogram-none-mse}
\end{figure}

\begin{figure}[h]
\centering

\begin{minipage}[t]{0.32\textwidth}
\centering
\includegraphics[width=\linewidth]{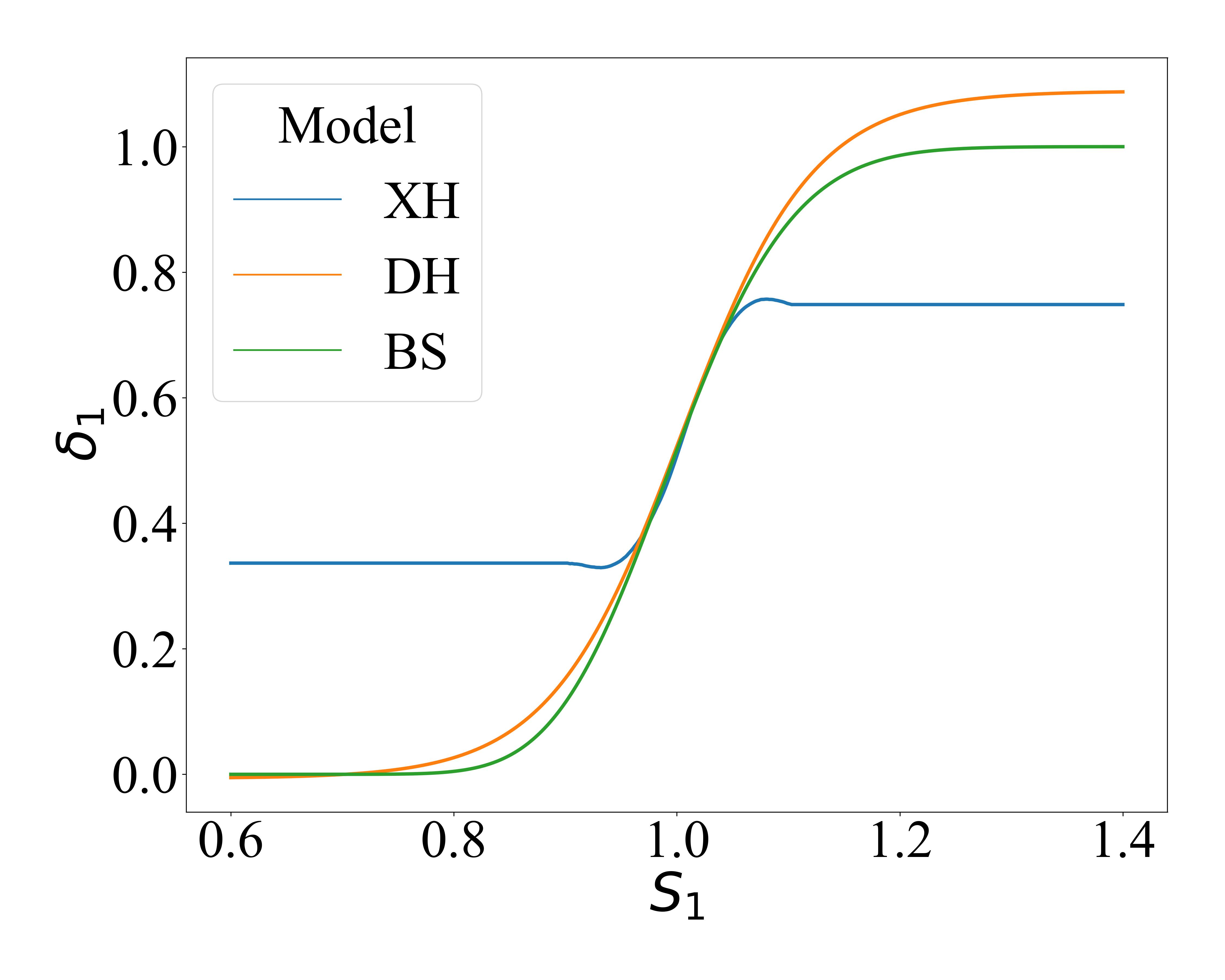}\\
(a) $k=1$.
\end{minipage}
\hfill
\begin{minipage}[t]{0.32\textwidth}
\centering
\includegraphics[width=\linewidth]{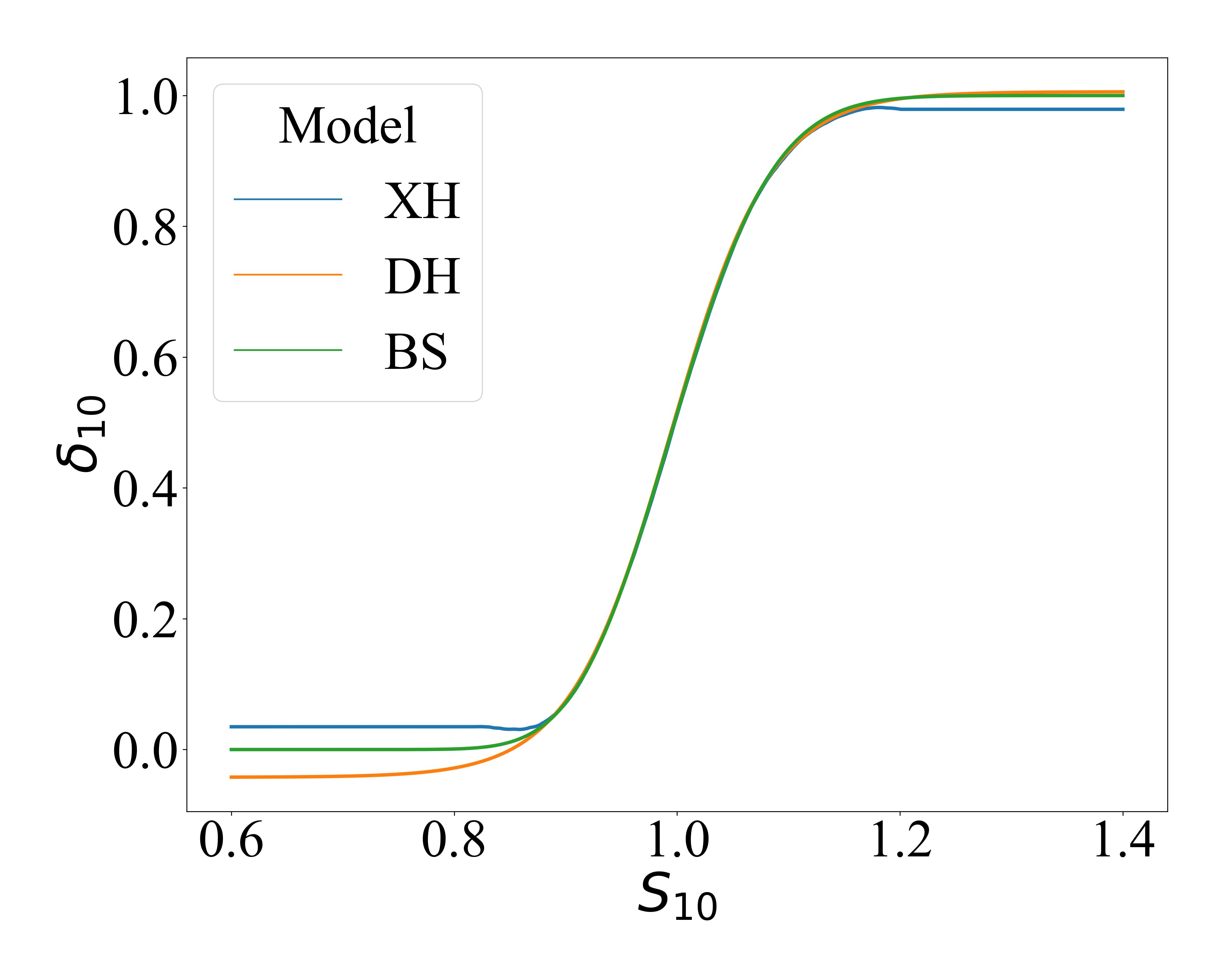}\\
(b) $k=10$.
\end{minipage}
\hfill
\begin{minipage}[t]{0.32\textwidth}
\centering
\includegraphics[width=\linewidth]{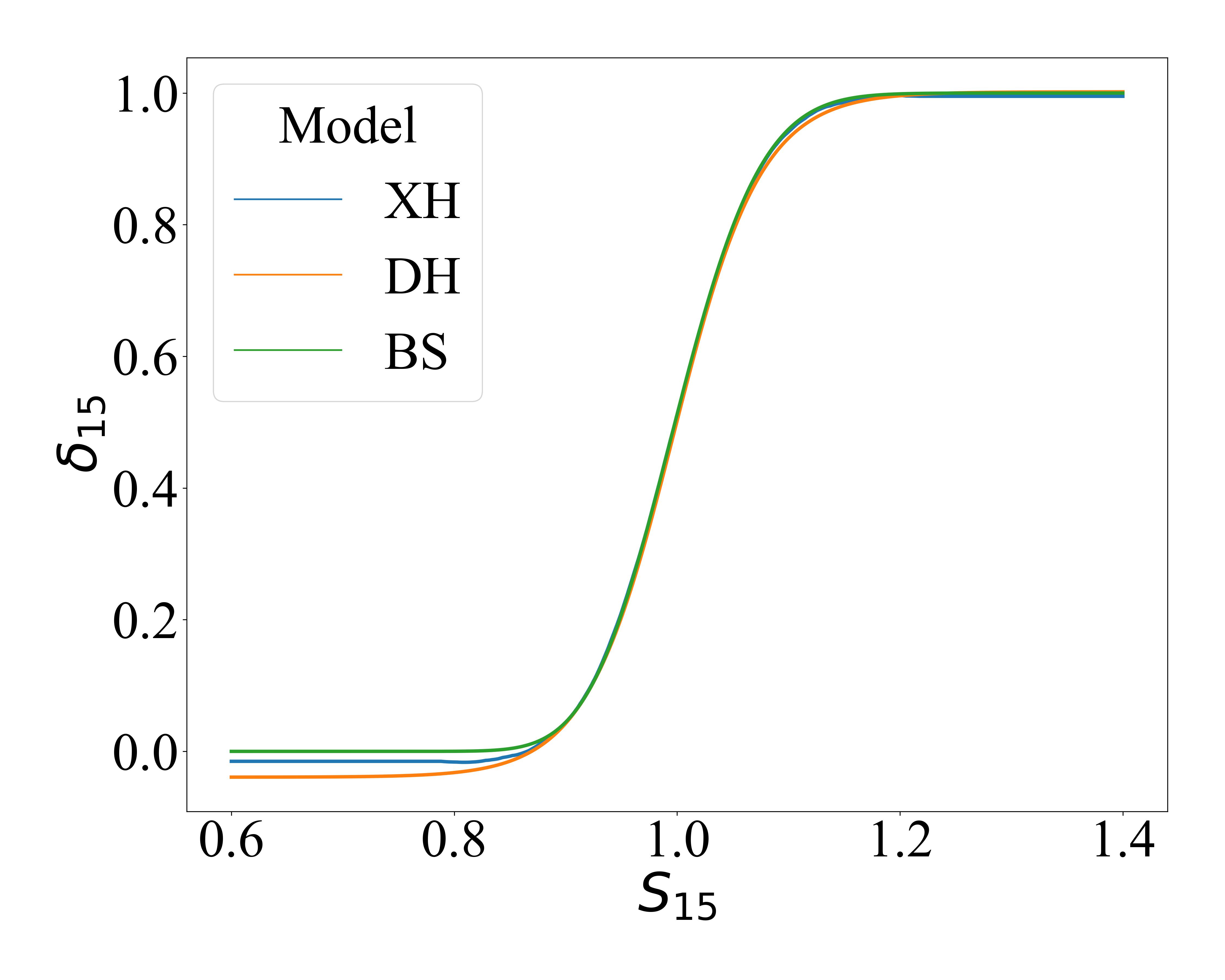}\\
(c) $k=15$.
\end{minipage}

\medskip

\begin{minipage}[t]{0.32\textwidth}
\centering
\includegraphics[width=\linewidth]{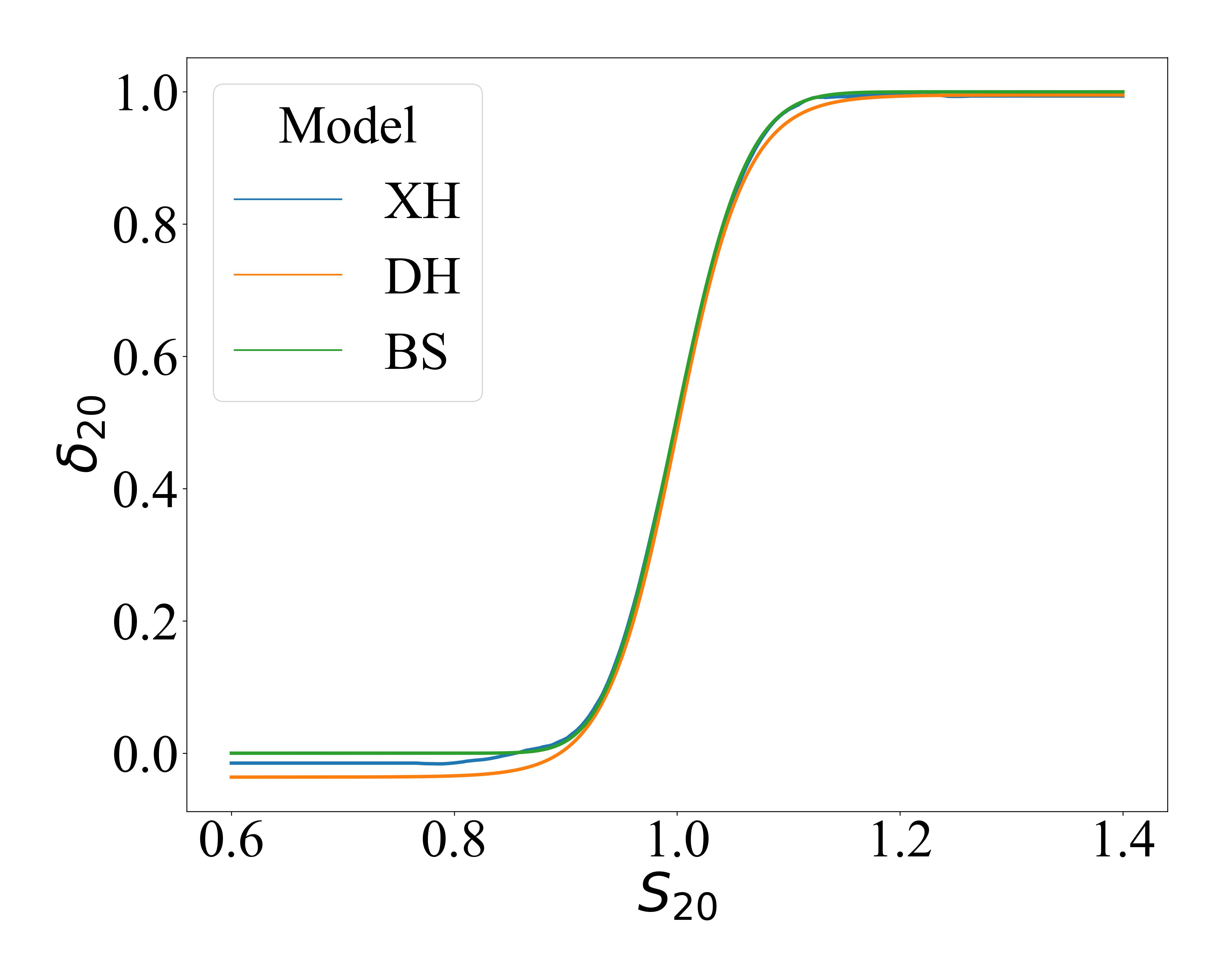}\\
(d) $k=20$.
\end{minipage}
\hfill
\begin{minipage}[t]{0.32\textwidth}
\centering
\includegraphics[width=\linewidth]{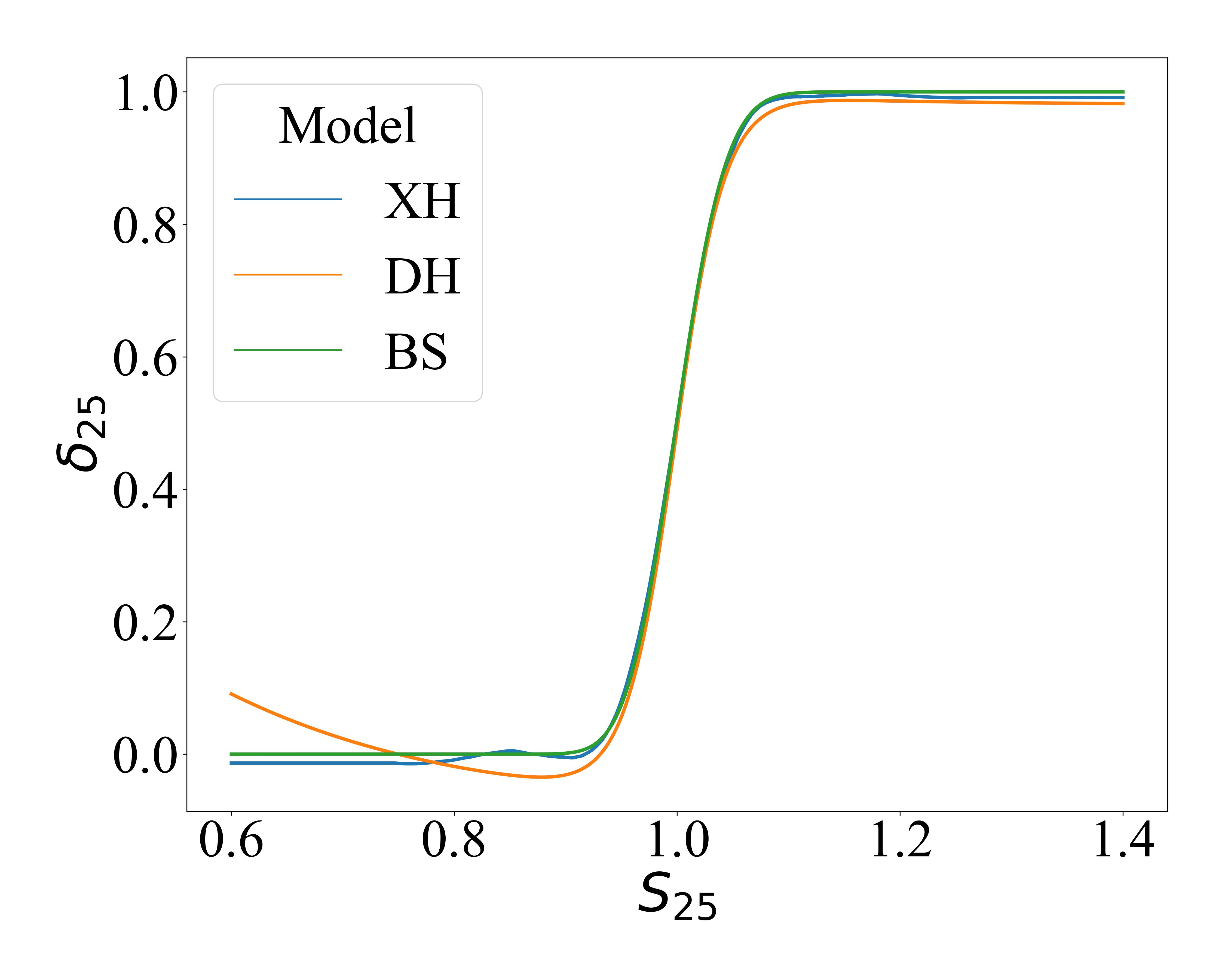}\\
(e) $k=25$.
\end{minipage}
\hfill
\begin{minipage}[t]{0.32\textwidth}
\centering
\includegraphics[width=\linewidth]{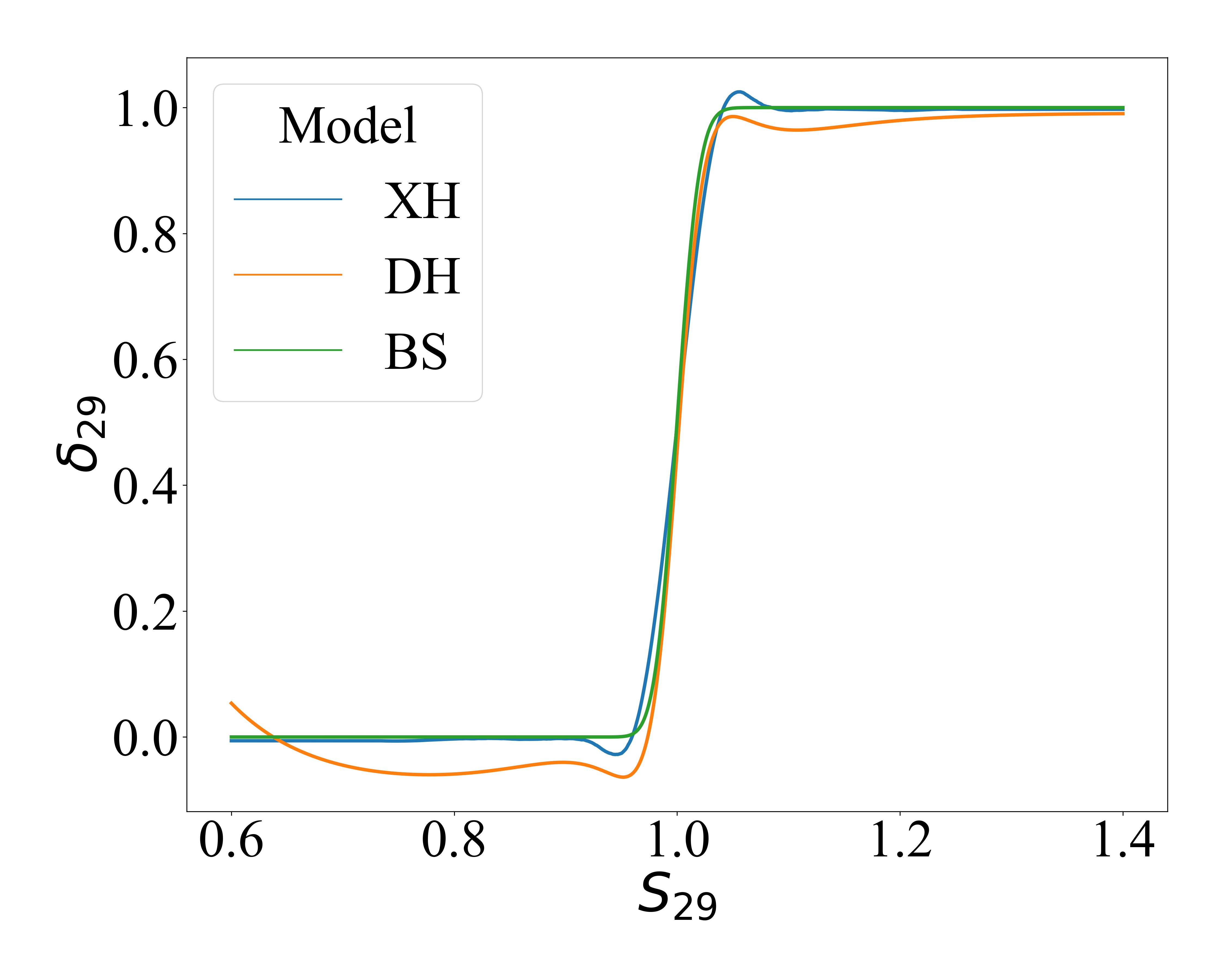}\\
(f) $k=29$.
\end{minipage}

\caption{Hedging strategies for ATM call option from times $k=1,10,15,20,25,29$ without transaction costs and using MSE loss function.}
\label{fig:strategy-noCost-mse-call}
\end{figure}

Table \ref{tab:results-none-mse} and the histogram in Figure \ref{fig:histogram-none-mse} present the X-Hedging results for the first baseline case, namely a European ATM call option, under the MSE loss function and in the absence of transaction costs. Figure \ref{fig:strategy-noCost-mse-call} illustrates the hedging strategies produced by the benchmark BS Hedging, Deep Hedging, and X Hedging approaches at selected time points. The qualitative insights from these plots are consistent with those observed in Figure \ref{fig:strategy-proportional-cvar-square-call} and discussed in detail in Section \ref{sec:performance_comparison}.

\begin{table}[h]
\caption{Mean, standard deviation and JS-divergence values for $\text{P\&L}_T$ distributions without transaction costs, using MSE loss function.}
\label{tab:results-none-mse}
\begin{tabular*}{\textwidth}{@{\extracolsep\fill}lrrrrr}
\toprule
&
\textbf{Mean} &
\textbf{Std} &
\textbf{JS(DH$\|\cdot$)} &
\textbf{JS(XH$\|\cdot$)} &
\textbf{Time (sec)} \\
\midrule
\textbf{BS} & -0.000001 & 0.005371 & 0.000273 & 0.000203 & 608.0 \\
\textbf{DH} & -0.000004 & 0.005423 & --- & 0.000277 & 66303.0 \\
\textbf{XH} &  0.000002 & 0.005505 & --- & --- & 45875.0 \\
\botrule
\end{tabular*}
\end{table}

\clearpage
\section{Results for hedging strategies with transaction costs and MSE loss function}
\label{app:baseCase2}
\setcounter{figure}{0}
\setcounter{table}{0}
Table \ref{tab:results-proportional-mse} and the histogram in Figure \ref{fig:histogram-proportional-mse} present the X-Hedging results for the second baseline case, namely a European ATM call option, under the MSE loss function and with transaction costs. Figure \ref{fig:strategy-proportional-mse}, analogous to Figure \ref{fig:strategy-noCost-mse-call}, illustrates the hedging strategies obtained from the benchmark BS-L Hedging, Deep Hedging, and X Hedging approaches at selected time points. The qualitative insights from these plots are consistent with those observed in Figure \ref{fig:strategy-proportional-cvar-square-call} and discussed in detail in Section \ref{sec:performance_comparison}.

\begin{table}[h]
\caption{Mean, standard deviation and JS-divergence values for $\text{P\&L}_T$ distributions with transaction costs and MSE loss function.}
\label{tab:results-proportional-mse}
\begin{tabular*}{\textwidth}{@{\extracolsep\fill}lrrrrr}
\toprule
&
\textbf{Mean} &
\textbf{Std} &
\textbf{JS(DH$\|\cdot$)} &
\textbf{JS(XH$\|\cdot$)} &
\textbf{Time (sec)} \\
\midrule
\textbf{BS-L} & -0.002251 & 0.005659 & 0.001324 & 0.001399 & 309.0 \\
\textbf{DH}   & -0.002202 & 0.005675 & --- & 0.000343 & 35514.0 \\
\textbf{XH}   & -0.002275 & 0.005763 & --- & --- & 21524.0 \\
\botrule
\end{tabular*}
\end{table}

\begin{figure}[h]
\centering
\includegraphics[width=0.7\textwidth]{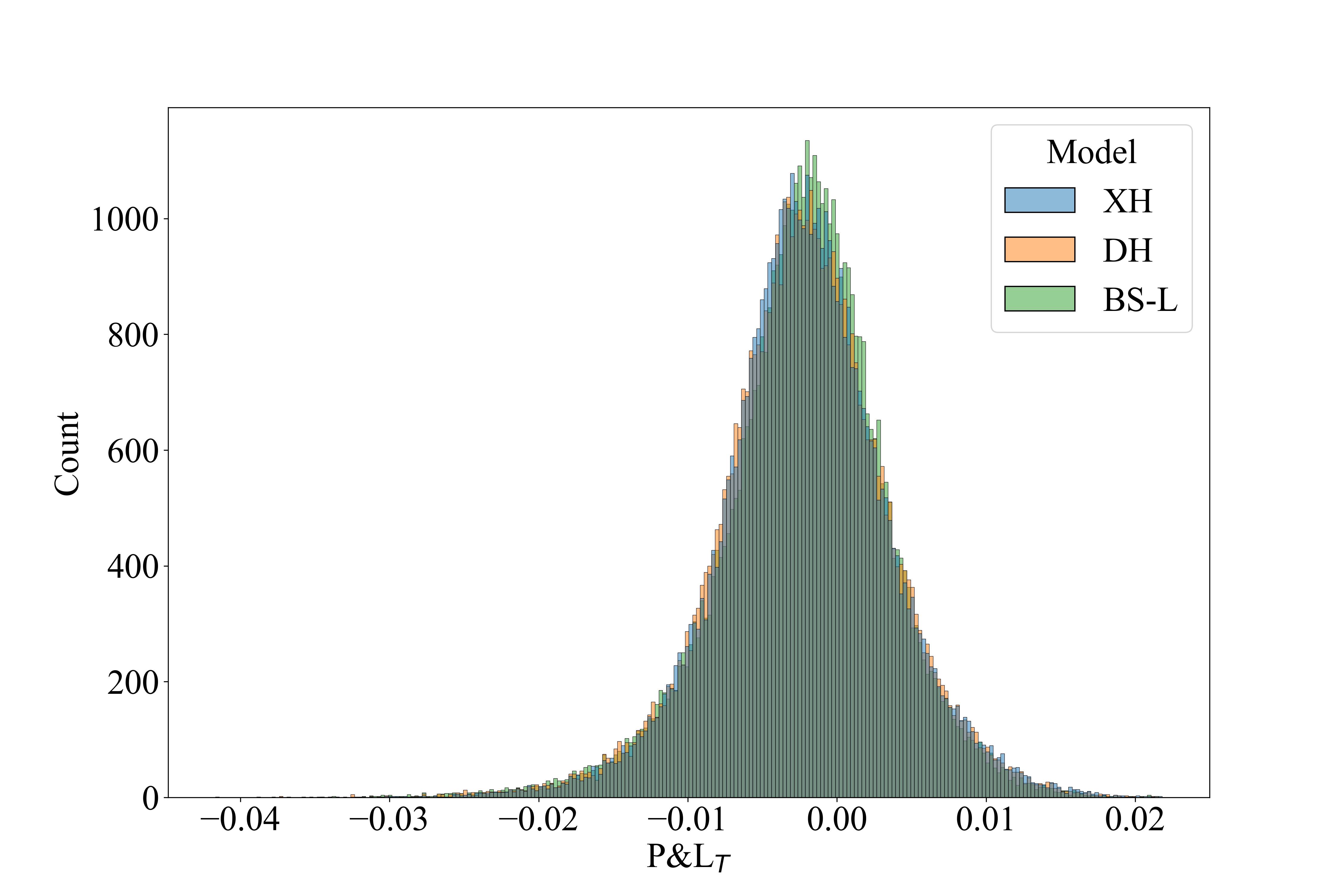}
\caption{Histograms of the $\text{P\&L}_T$ distributions with transaction costs and MSE loss function.}
\label{fig:histogram-proportional-mse}
\end{figure}

\begin{figure}[h]
\centering

\begin{minipage}[t]{0.32\textwidth}
\centering
\includegraphics[width=\linewidth]{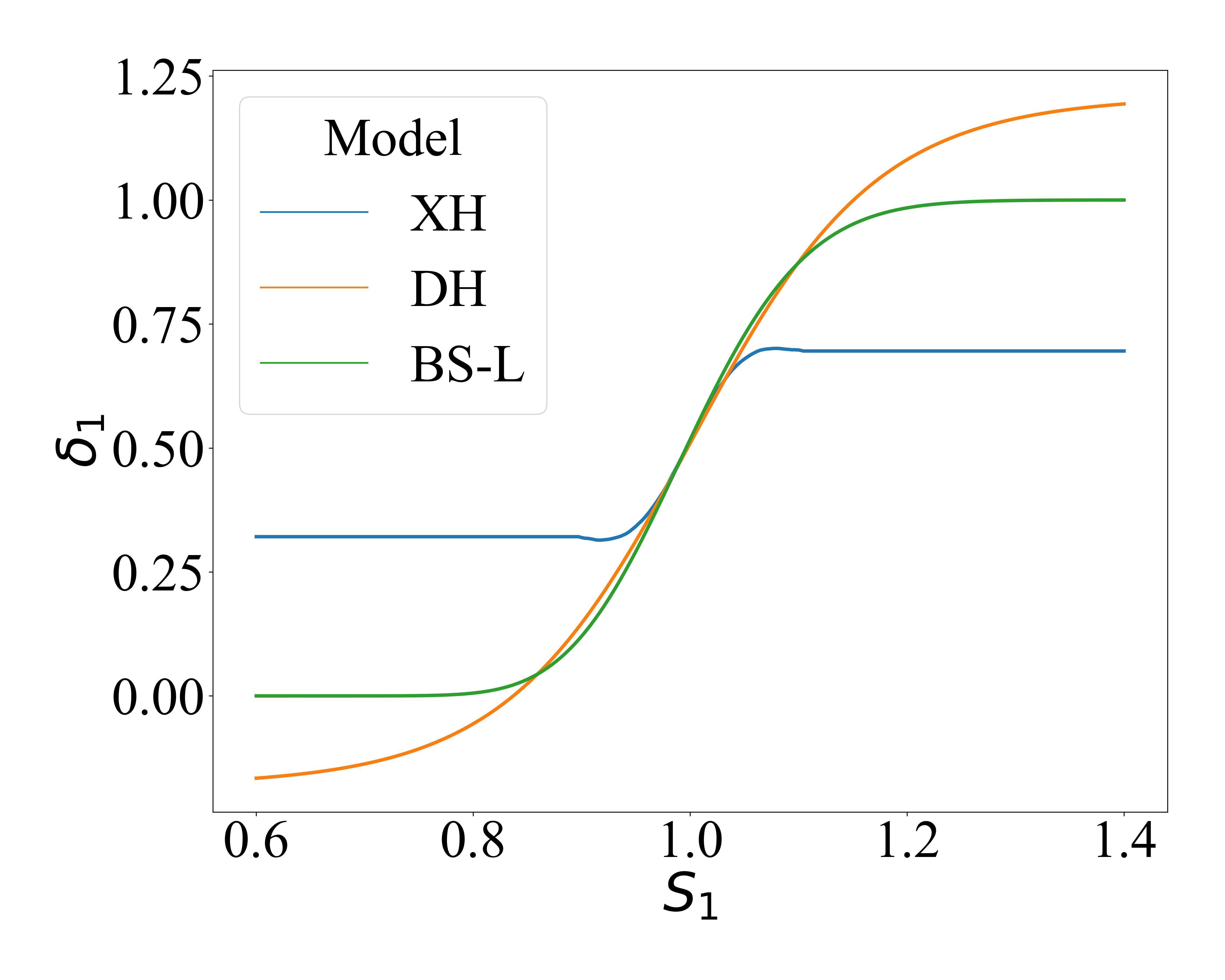}
\end{minipage}
\hfill
\begin{minipage}[t]{0.32\textwidth}
\centering
\includegraphics[width=\linewidth]{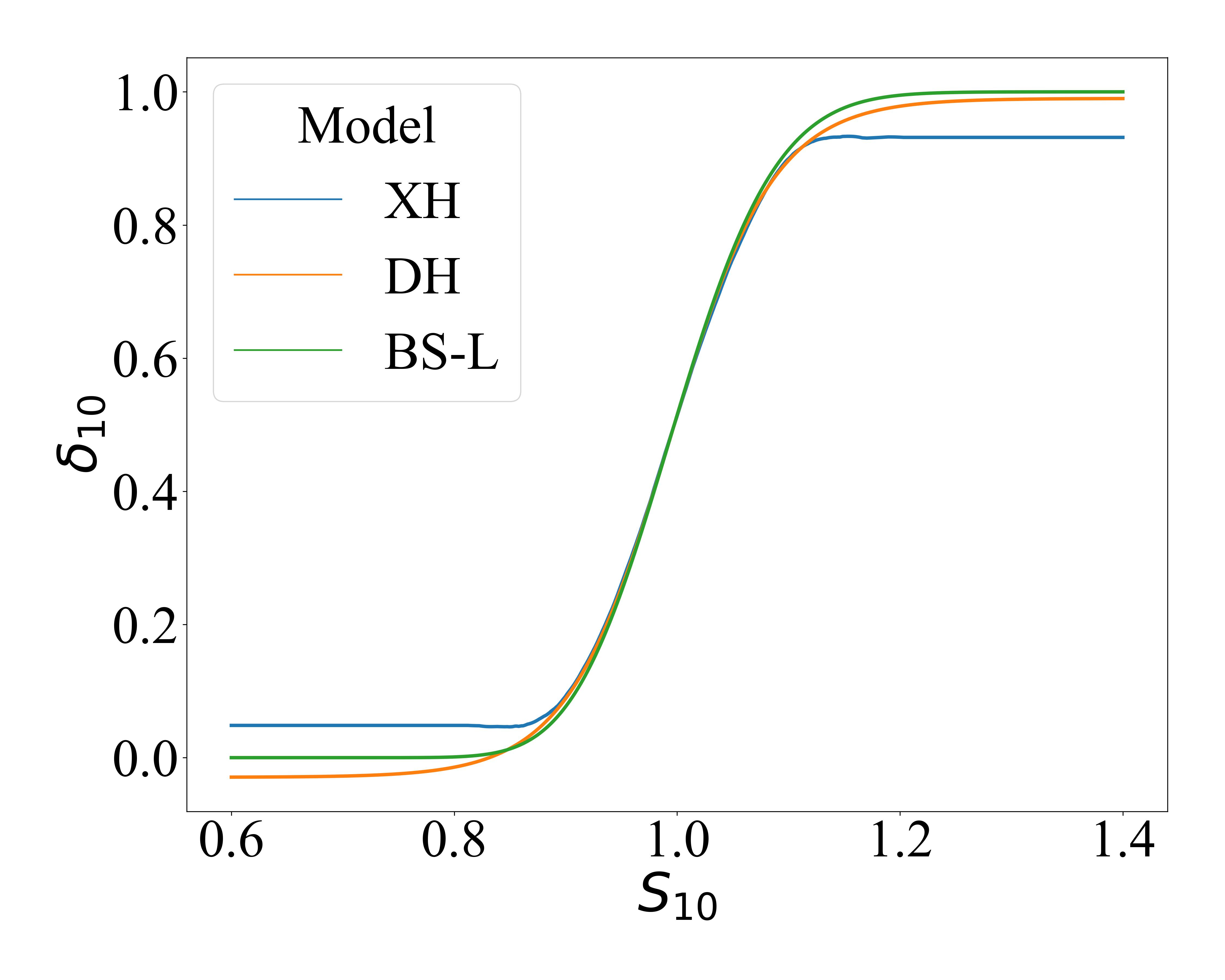}
\end{minipage}
\hfill
\begin{minipage}[t]{0.32\textwidth}
\centering
\includegraphics[width=\linewidth]{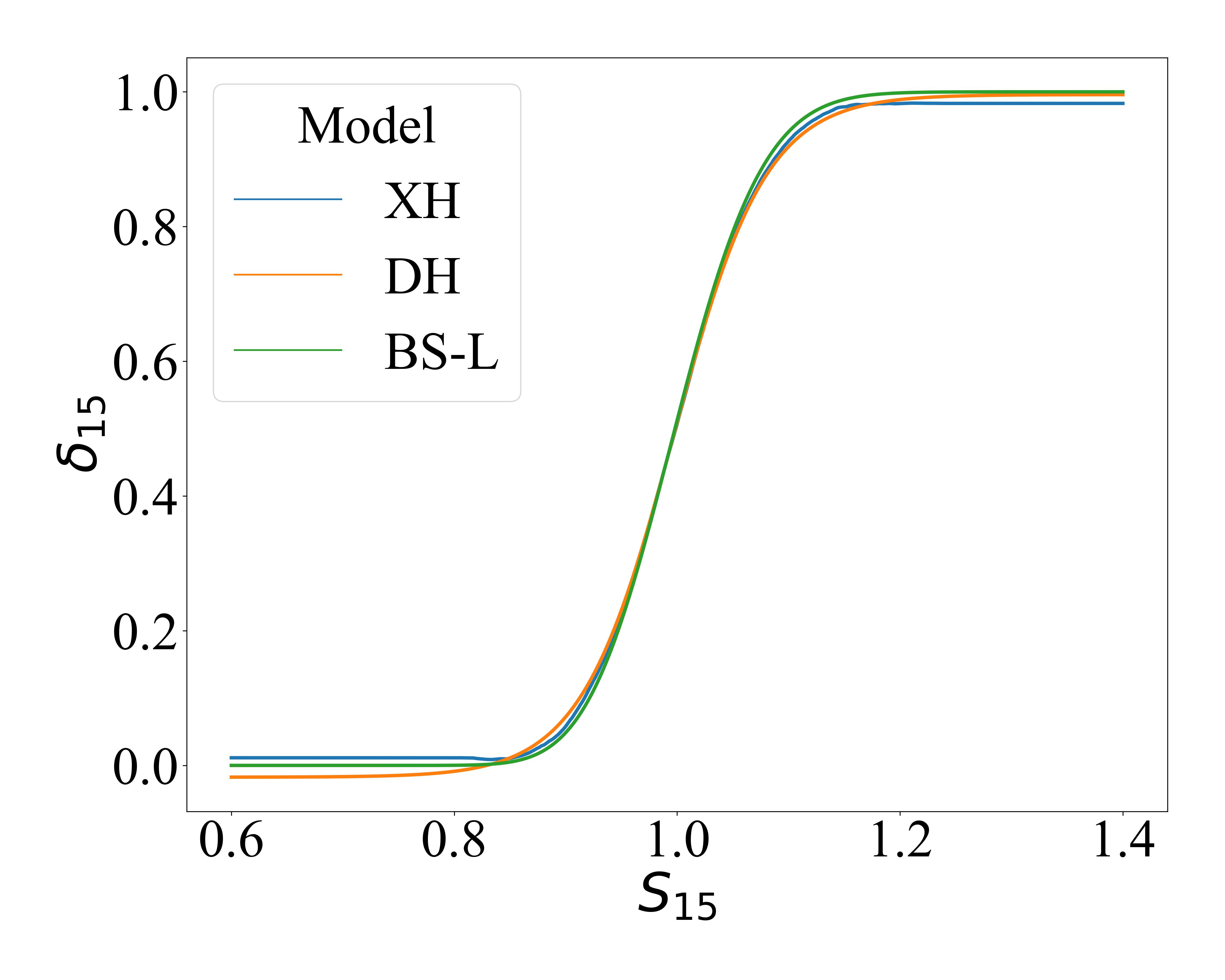}
\end{minipage}

\medskip

\begin{minipage}[t]{0.32\textwidth}
\centering
\includegraphics[width=\linewidth]{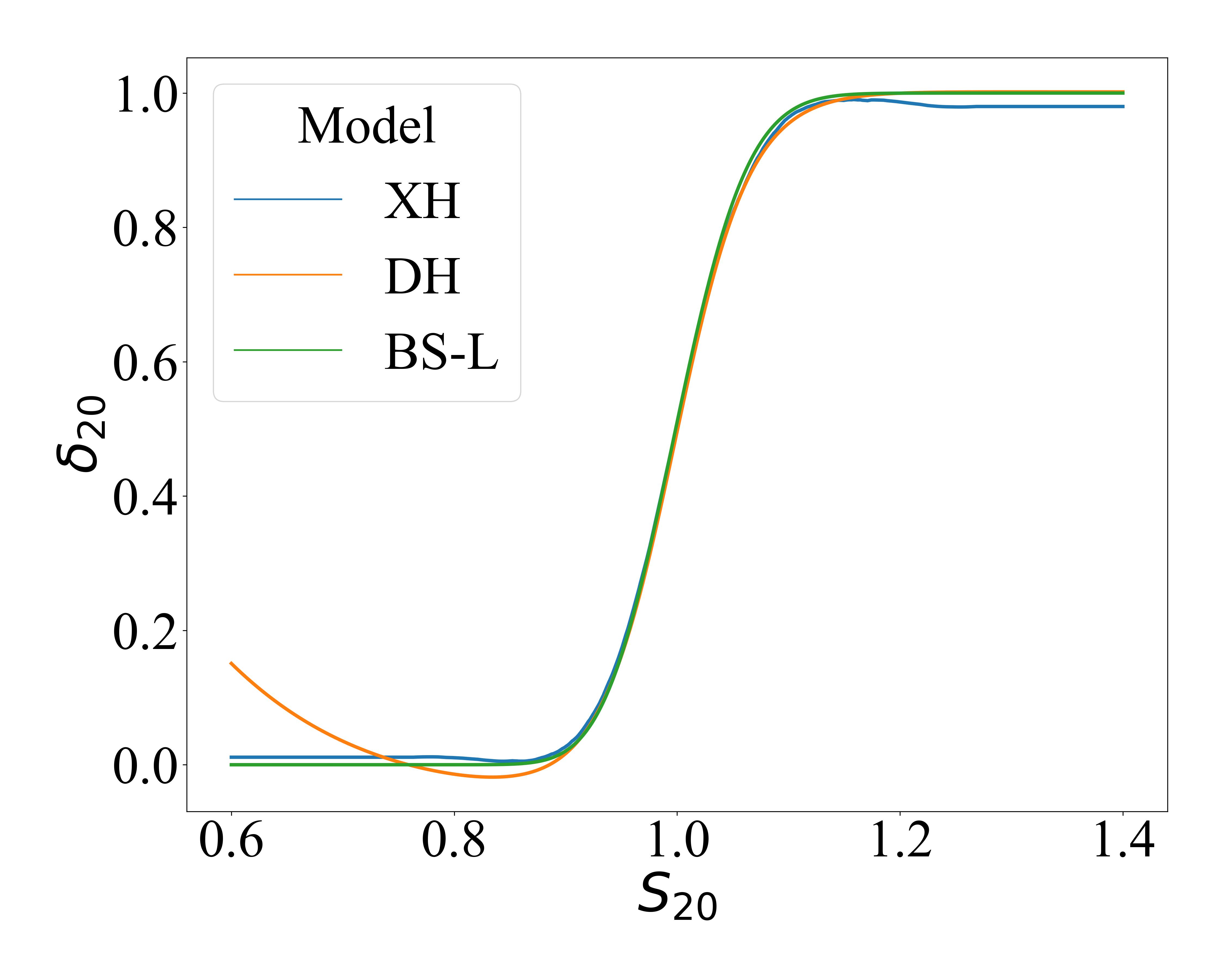}
\end{minipage}
\hfill
\begin{minipage}[t]{0.32\textwidth}
\centering
\includegraphics[width=\linewidth]{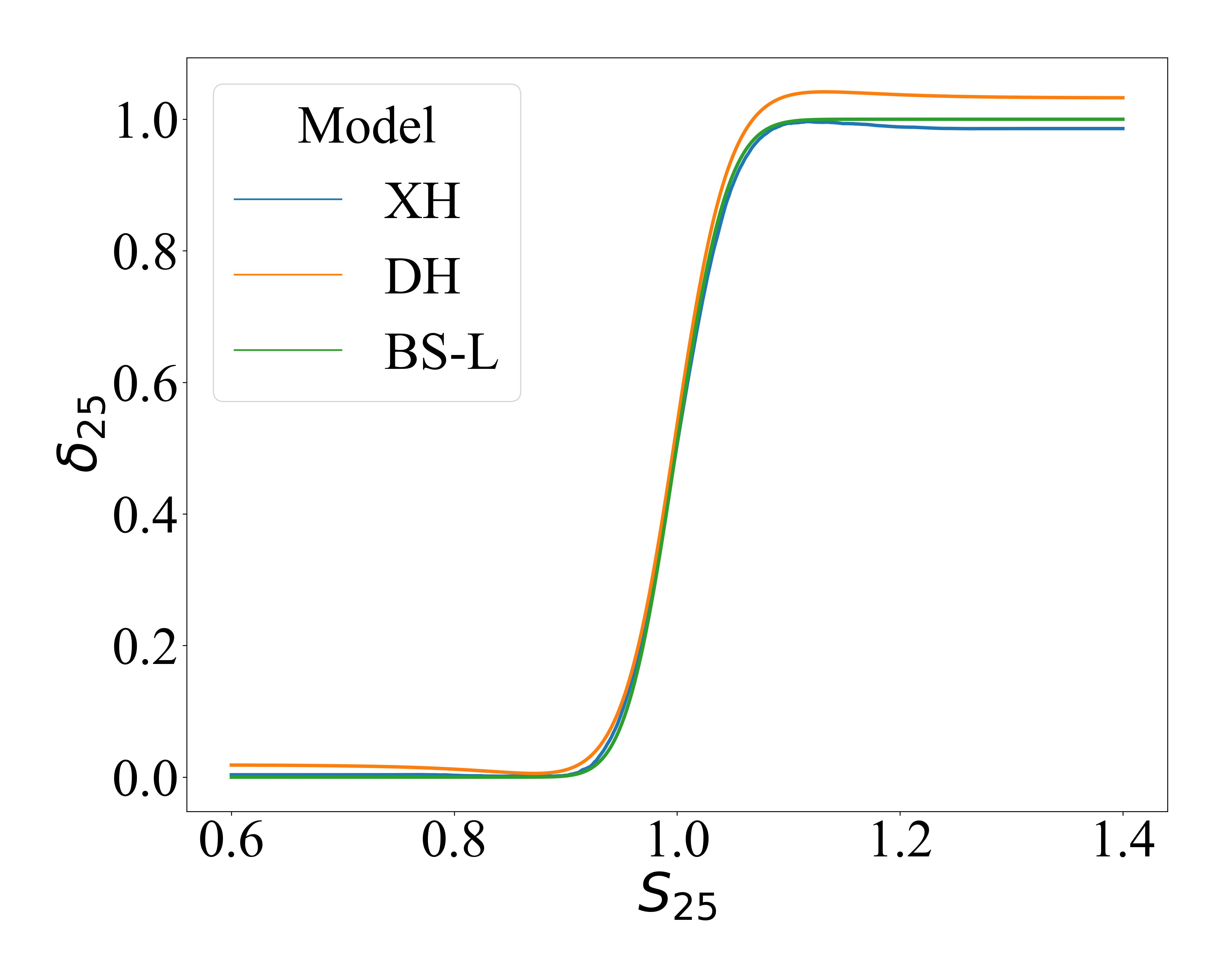}
\end{minipage}
\hfill
\begin{minipage}[t]{0.32\textwidth}
\centering
\includegraphics[width=\linewidth]{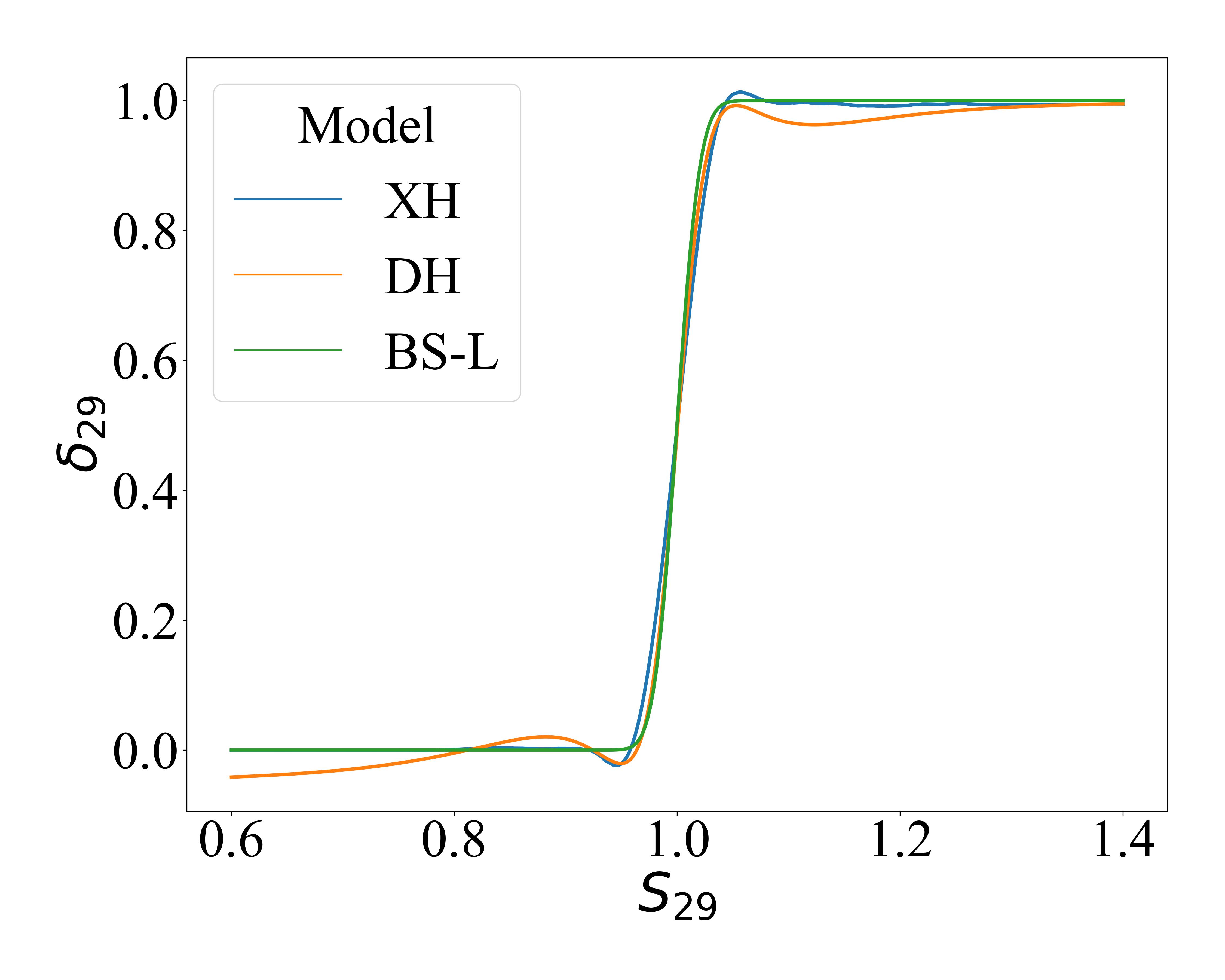}
\end{minipage}

\caption{Hedging strategies for ATM call option from times
$k=1,10,15,20,25,29$ with transaction costs and using MSE loss function.}
\label{fig:strategy-proportional-mse}
\end{figure}

\clearpage
\section{Results for hedging strategies with transaction costs and CTM$_2$ loss function for various moneyness}
\label{app:allmoneyness}
\setcounter{figure}{0}
\setcounter{table}{0}

We examine in-the-money (ITM) and out-of-the-money (OTM) call and put options. The strike price is set to two values, $K=0.9$ and $K=1.1$, corresponding to the ITM and OTM cases for call options and the OTM and ITM cases for put options, respectively. The statistical properties of the hedging strategies generated by X Hedging, Deep Hedging, and BS-L Hedging are summarized in Tables \ref{tab:results-call-put-itm} (ITM options) and \ref{tab:results-call-put-otm} (OTM options).

\begin{table}[h]
\caption{Mean, standard deviation (Std), and JS-divergence values for $\text{P\&L}_T$ distributions under transaction costs and the CTM$_2$ loss function for ITM options.}
\label{tab:results-call-put-itm}
\begin{tabular*}{\textwidth}{@{\extracolsep\fill}llrrrrr}
\toprule
& &
\textbf{Mean} &
\textbf{Std} &
\textbf{JS(DH$\|\cdot$)} &
\textbf{JS(XH$\|\cdot$)} &
\textbf{Time (sec)} \\
\midrule
& \textbf{BS-L} & -0.001677 & 0.003397 & 0.03312  & 0.032953 & 565.0 \\
\textbf{Call} & \textbf{DH} & -0.001689 & 0.003618 & --- & 0.001417 & 58007.0 \\
& \textbf{XH} & -0.001756 & 0.003664 & --- & --- & 30911.0 \\
\midrule
& \textbf{BS-L} & -0.002227 & 0.005642 & 0.004922 & 0.006386 & 315.0 \\
\textbf{Put} & \textbf{DH} & -0.002130 & 0.005845 & --- & 0.000344 & 24532.0 \\
& \textbf{XH} & -0.002164 & 0.005864 & --- & --- & 26150.0 \\
\botrule
\end{tabular*}
\end{table}

\begin{table}[h]
\caption{Mean, standard deviation (Std), and JS-divergence values for $\text{P\&L}_T$ distributions under transaction costs and CTM$_2$ loss function for OTM options.}
\label{tab:results-call-put-otm}
\begin{tabular*}{\textwidth}{@{\extracolsep\fill}llrrrrr}
\toprule
& &
\textbf{Mean} &
\textbf{Std} &
\textbf{JS(DH$\|\cdot$)} &
\textbf{JS(XH$\|\cdot$)} &
\textbf{Time (sec)} \\
\midrule
& \textbf{BS-L} & -0.001142 & 0.004263 & 0.041733 & 0.029112 & 622.0 \\
\textbf{Call} & \textbf{DH} & -0.001115 & 0.004475 & --- & 0.003673 & 49020.0 \\
& \textbf{XH} & -0.001233 & 0.004558 & --- & --- & 35003.0 \\
\midrule
& \textbf{BS-L} & -0.000907 & 0.003400 & 0.043407 & 0.031675 & 447.0 \\
\textbf{Put} & \textbf{DH} & -0.001043 & 0.003964 & --- & 0.004696 & 43102.0 \\
& \textbf{XH} & -0.000995 & 0.003695 & --- & --- & 44527.0 \\
\botrule
\end{tabular*}
\end{table}

\end{appendices}


\bibliography{Xhedging_IJACM}

\end{document}